OXFORD

Systems Biology

# MolPLA: A Molecular Pretraining Framework for Learning Cores, R-Groups and their Linker Joints

**Mogan Gim [1,†], Jueon Park [1†], Soyon Park [1], Sanghoon Lee [1,2,], Seungheun Baek [1], Junhyun Lee [1], Ngoc-Quang Nguyen [1], and Jaewoo Kang [1,2,∗]**

[1]Department of Computer Science, Korea University, Seoul 02841, Republic of Korea and
[2]AIGEN Sciences, Seoul 04778, Republic of Korea
[†]Equal Contributors.

[∗]To whom correspondence should be addressed.

## Abstract

**Motivation:** Molecular core structures and R-groups are essential concepts in drug development. Integration of these concepts with conventional graph pre-training approaches can promote deeper understanding in molecules. We propose MolPLA, a novel pre-training framework that employs masked graph contrastive learning in understanding the underlying decomposable parts in molecules that implicate their core structure and peripheral R-groups. Furthermore, we formulate an additional framework that grants MolPLA the ability to help chemists find replaceable R-groups in lead optimization scenarios.
**Results:** Experimental results on molecular property prediction show that MolPLA exhibits predictability comparable to current state-of-the-art models. Qualitative analysis implicate that MolPLA is capable of distinguishing core and R-group sub-structures, identifying decomposable regions in molecules and contributing to lead optimization scenarios by rationally suggesting R-group replacements given various query core templates.
**Availability:** The code implementation for MolPLA and its pre-trained model checkpoint is available at https://github.com/dmis-lab/MolPLA
**Contact:** kangj@korea.ac.kr
**Supplementary information:** Supplementary data are available at *Bioinformatics* online.

## 1 Introduction

Molecular representation learning is essential in success of AI-driven drug discovery applications (Gim *et al.*, 2023; Nguyen *et al.*, 2024). Various deep neural network architectures and representation methods have been proposed to accurately predict molecular properties or effectively generate novel compounds. Particularly, graph-based representations have become popular choices which naturally motivated the use of Graph Neural Networks (GNNs) as they encode the molecule’s characteristics based on its topological features. Nevertheless, current downstream applications have struggled against performance bottlenecks due to complexity of molecules and lack of their labeled data.

To address these issues, researchers have proposed large-scale pre-training frameworks that exploit the abundance of molecules in self-supervised learning settings. They are renowned for helping representation models effectively characterize chemical compounds and boosting their performance in various downstream tasks. Masked graph modeling (MGM) frameworks first mask partial parts of the given molecule then train the model to recover the missing information using its surrounding chemical context (Rong *et al.*, 2020; Zhang *et al.*, 2021; Liu *et al.*, 2022). Graph contrastive learning (GCL) frameworks on the other hand, modulate the model’s inherent high-dimensional embedding space by aligning two different views of the same molecule (Wang *et al.*, 2022b,a; Wu *et al.*, 2023).

One important aspect in GCL frameworks is learning global (e.g. bio-chemical properties) and local structural features (e.g. functional groups) of molecular graphs. Recent works have sought ways to incorporate cheminformatics-related concepts in GCL, particularly molecular fragments (e.g. building blocks, motifs or functional groups). Applications of fragment-based drug discovery inspired many researchers to incorporate such concepts which exhibited significant improvements in property prediction tasks as decomposed fragments of the same molecule contribute differently to its various properties (Zhang *et al.*, 2020).

Meanwhile, scaffolds (i.e. cores) and R-groups (i.e. substituents, side chains) which are the backbone and its peripheral parts of molecules respectively, are popularly employed concepts in cheminformatics. Just as important as fragments, scaffolds have been widely employed in drug discovery applications such as molecular generation (Li *et al.*, 2019). In medicinal chemistry, they determine the overall structure and bioactivity properties of molecules when engaged in interaction with target proteins (Li *et al.*, 2019). For instance, scaffold-constrained generation methods (i.e. scaffold decoration) start with a hard scaffold and optimize its properties by adding other R-groups to it (Maziarz *et al.*, 2021).

R-groups are defined as peripheral components attached to the molecule's core structure. Their relationship with molecular core plays a key role in bioisosterism which refers to modification of lead compounds with the objective of retaining, improving or altering its biochemical characteristics (Papadatos and Brown, 2013). Medicinal chemists determine which structural perturbations (atom or R-group substitution) to apply to which parts of the molecule to designate as decomposable site. For the rest of the paper, we denote these sites as *linker joints* that comprise linker nodes and edges.

In this work, we introduce *MolPLA* , a novel pre-training framework that builds molecular graph representations based on the concepts of cores and R-groups. Our main focus in this study is to explore the advantages of incorporating these concepts in GCL. To the best of our knowledge, none of the recent works have employed these concepts in pre-training methods.

We designed a novel molecular graph decomposition method by leveraging Naveja *et al.*'s approach for manually identifying putative cores of a molecule and masked the linker joints of its decomposed parts. The proposed method builds various molecular view pairs that comprise the original molecule with its decomposed form of core and R-groups, and is used for construction of pre-training dataset originated from GEOM (Axelrod and Gomez-Bombarelli, 2022). We devised a masked graph contrastive learning framework (MGCL) that imposes two training objectives which are GCL between original and decomposed molecule and MGM through maximizing agreement between the original linker nodes and their corresponding masked ones used in decomposition.

We extended *MolPLA* 's functionality by devising R-group retrieval (RGR) framework which leads to task formulation where the model's objective is to retrieve the correct R-groups that were previously decoupled in its decomposition process. The intuition is to exploit *MolPLA* 's understanding in molecular core, R-groups and their linker joints in contributing to practical drug discovery tasks such as scaffold-constrained lead optimization which involves reasonable retrieval and substitution of R-groups in a reference molecule's core template.

Overall, *MolPLA* consists of the MGCL and RGR framework. The former helps the model understand the inherent decomposable parts of a molecule that make a structural distinction between its core and R-groups. The latter enables it to retrieve R-groups that are suitable to the query core template's masked linker node that acts as a re-attachment point.

Experimental results on nine molecular property prediction tasks compared with other pre-training baselines and ablations demonstrate the effectiveness of the proposed MGCL framework in *MolPLA* . Moreover, additional results on the R-group retrieval task reveal that *MolPLA* shows reliable performance in not only the test partition of our pre-training dataset but also external dataset. Qualitative analysis on its learned node representations demonstrates *MolPLA* 's understanding in decomposable regions of molecules. Lead optimization results demonstrate its potential in helping researchers design novel molecules with improved drug-likeness properties.

| Symbol | Description |
|---|---|
| **M** | Molecule |
| **D** | Decomposed Molecule |
| **C** | Putative Core |
| **Q** | Query Template |
| **R** | R-Group |
| $\mathcal{G}$ | Molecular Graph or Sub-graph |
| $\mathbb{G}$ | Set of Molecular Graphs or Sub-graphs |
| $\mathcal{G}^{\circ}$ | $\mathcal{G}$ with at least one masked Linker Joint |
| $\mathcal{D}$ | Data instance in pre-training dataset |

Table 1. Table of notations.

## 2 Materials and Methods

### 2.1 Dataset

#### 2.1.1 Putative Cores and R-Groups

We pre-processed 304,466 drug-like compounds originated from GEOM (Axelrod and Gomez-Bombarelli, 2022) to construct a pre-training dataset for *MolPLA* . Our central approach for implementing *MolPLA* lies in the concept of decomposing molecules into its identified core-like sub-structure and R-groups. One of the most commonly used scaffold concepts is the Bemis-Murcko scaffolds (Bemis and Murcko, 1996) of which its key limitations are inconsideration of retrosynthesis rules and neglection of ring-like substituents (Naveja *et al.*, 2019). To overcome these limitations, an improved scaffold concept called analog series-based scaffolds (ASBS) was introduced by Stumpfe *et al.*. Subsequently, Naveja *et al.* proposed a novel scaffold concept called putative cores which retains the advantages of ASBS but adds flexibility and diversity in identifying the scaffold-like substructures of a molecule. The first two concepts follow the "single molecule-single scaffold" paradigm while putative cores extend it to "single molecule-multiple scaffolds" (Naveja *et al.*, 2019). An illustrated comparison between two molecular decomposition methods using Murcko scaffolds and Naveja's putative cores is available in the supplementary material **S2**.

Applying this concept to GCL will not only help the model understand the meaningful structural features of the molecule but also benefit from data augmentation effects. Our novel molecular graph decomposition method utilizes Naveja *et al.*'s putative core framework. For each molecule contained in the GEOM dataset, we first identify its multiple number of putative cores. In perspective of graphs, a molecular graph $\mathcal{G}_{\mathbf{M}}$ can have one or many putative core sub-graphs $\mathcal{G}_{\mathbf{C}}$.

For each putative core of molecule, we identified R-groups as multiple peripheral substructures bonded to it. Since there may exist multiple cores for a single molecule, each of their corresponding set of R-groups may differ as well. By applying graph-based definition, we denote the sub-graph representations of R-groups as $\mathcal{G}_{\mathbf{R}}$. All molecules, putative cores and R-groups are represented as molecular graphs or sub-graphs which enables the formulation of our proposed molecular graph decomposition method.

As explained above, multiple putative cores were identified in each molecule which resulted in 1,231,364 data instances of ($\mathcal{G}_{\mathbf{M}}$, $\mathcal{G}_{\mathbf{C}}$). As some large molecules may have excessive number of identified putative cores, we excluded data tuples whose molecule has more than 10 cores since the statistics revealed 11 cores represent the 99th percentile of the dataset. As a result, the number of data instances consisting a molecule with its identified putative cores at this pre-processing step is 1,196,157.

#### 2.1.2 Proposed Molecular Graph Decomposition

A traditional decomposition method would involve bond cleavage between two connecting atoms. We instead propose our novel method based on one linker atom shared by putative core and R-group. For each pair of identified

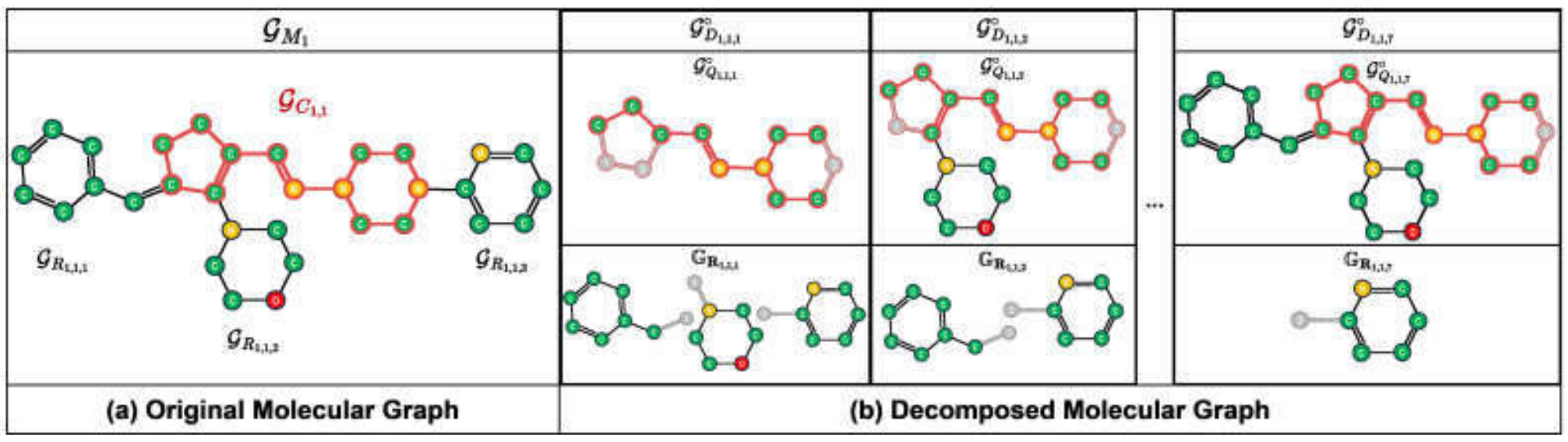

Fig. 1: Schematic illustration for our proposed molecular graph decomposition method. Given a molecule represented as its graph structure $\mathcal{G}_{\mathbf{M}_1}$ and one of its putative cores $\mathcal{G}_{\mathbf{C}_{1,1}}$ identified by Naveja *et al.*'s framework, multiple decomposition results can be obtained. Since the number of R-groups is 3, the total number of decomposition results including decoupling all R-groups from the putative core, is 7. Additional figures are available in the supplementary material **S4**, **S5**

putative core and R-groups connected by a linker node, we *decouple* a subset of R-groups from the core while preserving the shared linker nodes. As there are multiple R-groups for each molecule with its putative core, multiple decomposition results can be obtained. For example in Figure 1, if a molecular graph $\mathcal{G}_{\mathbf{M}}$ with identified putative core $\mathcal{G}_{\mathbf{C}}$ has three R-groups ($\{\mathcal{G}_{\mathbf{R}_1}, \mathcal{G}_{\mathbf{R}_2}, \mathcal{G}_{\mathbf{R}_3}\}$), then seven decomposition results are obtained including all R-groups being decoupled.

The molecular graph decomposition method used to build data instance $\mathcal{D}_{i,j,k}$ is mathematically defined as follows,

$$\mathcal{D}_{i,j,k} = (\mathcal{G}_{\mathbf{M}_i}, \mathcal{G}_{\mathbf{D}_{i,j,k}}) \tag{1}$$

$$\mathcal{G}_{\mathbf{D}_{i,j,k}} = (\mathcal{G}_{\mathbf{Q}_{i,j,k}}, \mathbb{G}_{\mathbf{R}_{i,j,k}}) \tag{2}$$

$$\mathcal{G}_{\mathbf{Q}_{i,j,k}} = \mathcal{G}_{\mathbf{C}_{i,j}} \cup (\mathbb{G}_{\mathbf{R}_{i,j}} - \mathbb{G}_{\mathbf{R}_{i,j,k}}) \tag{3}$$

where $\mathcal{G}_{\mathbf{D}_{i,j,k}}$ refers to the $i$th molecule with its identified $j$th putative core $\mathcal{G}_{\mathbf{C}_{i,j}}$ and union set of R-groups $\mathbb{G}_{\mathbf{R}_{i,j}}$ being decomposed into $\mathcal{G}_{\mathbf{Q}_{i,j,k}}$ and $\mathbb{G}_{\mathbf{R}_{i,j,k}}$. $\mathbb{G}_{\mathbf{R}_{i,j,k}}$ is a $k$th subset of R-group sub-graphs decoupled from the original molecule while $\mathcal{G}_{\mathbf{Q}_{i,j,k}}$ is its remaining graph including core sub-structure ($\mathcal{G}_{\mathbf{C}_{i,j}}$). For the rest of the paper, we denote $\mathcal{G}_{\mathbf{Q}_{i,j,k}}$ as the molecular query template.

Previous graph self-supervised learning approaches empirically demonstrated the advantages of MGM techniques in building more robust representation models. Inspired by them, we masked the node and edge attributes of decoupled linker joints contained in each decomposed molecular graph $\mathcal{G}_{\mathbf{D}_{i,j,k}}$. For example in Figure 1-(b)'s $\mathcal{G}_{\mathcal{D}_{1,1,2}}$ where $\mathcal{G}_{\mathbf{R}_1}$ and $\mathcal{G}_{\mathbf{R}_3}$ are being decoupled from putative core $\mathcal{G}_{\mathbf{C}}$, the graph decomposition applies node and edge attribute masking to its linker joints except the one where $\mathcal{G}_{\mathbf{R}_2}$ is still intact.

The final result of the proposed molecular decomposition method is mathematically defined as follows,

$$\mathcal{D}'_{i,j,k} = (\mathcal{G}_{\mathbf{M}_i}, \mathcal{G}^{\circ}_{\mathbf{D}_{i,j,k}}) \tag{4}$$

The masked linker joints in our graph decomposition method are not only expected to enhance our model's graph embedding space of molecules but also offer flexibility in perspective of lead optimization. By decoupling and masking the linker joints between putative cores and R-groups, we can now liberate *MolPLA* 's R-group retrieval process from the predefined chemical restrictions that were contained in the original molecular graph structure. That is, any R-group retrieved by *MolPLA* can be used to optimize lead compounds despite their original linker joints having different chemical attributes especially atom symbols and bond types.

### 2.1.3 Additional Dataset Pre-processing

The proposed molecular decomposition resulted in a total of 62,196 masked R-groups available in the dataset. We observed that their occurrence count in data instances shows a skewed distribution of less informative R-groups such as -OH. To mitigate this, we first designated those with more than 99.99th percentile as common R-groups. We then removed data instances ($\mathcal{G}^{\circ}_{\mathbf{D}_{i,j,k}}$) whose over half of their decoupled masked R-groups ($\mathbb{G}_{\mathbf{R}_{i,j,k}}$) are common ones.

Furthermore, for each of the masked R-groups available in the dataset, we added binary feature vectors where each bit position indicates the presence of a particular functional group. We denote them as R-Group Condition Vectors as they will guide *MolPLA* 's R-group retrieval task. Our main rationale relates to a situation where medicinal chemists initially set the structural requirements of a molecular sub-structure when searching for new synthesis idea.

The final number of data instances in our pre-training dataset is 1,054,787 built from 304,466 drug-like molecules. Detailed statistics and illustrated pre-processing steps of this dataset can be found in the supplementary material **S1 S3**.

## 2.2 Model

*MolPLA* consists of two frameworks which are Masked Graph Contrastive Learning (MGCL) and R-Group Retrieval (RGR) Framework. In the MGCL framework, *MolPLA* learns both local and global characteristics of a molecule based on its decomposable structure-related features. In the RGR framework, *MolPLA* learns how to retrieve an attachable R-group to the decomposed molecule mapped by their respective linker joints. Figure 2 illustrates an overview of *MolPLA* .

### 2.2.1 Masked Graph Contrastive Learning Framework

The MGCL framework takes an input molecule to build two contrastive graph projections in molecular graph embedding space and guides *MolPLA* to maximize their alignment within the space. It comprises weight-sharing graph encoders $f_\theta$, readout layers, weight-sharing graph projection heads $g_\theta$ and node projection heads $g_\kappa$.

The input molecule is first converted into two distinct graph-based views which are its original $\mathcal{G}_{\mathbf{M}}$ and decomposed $\mathcal{G}_{\mathbf{D}}$ molecular graph. The two views are converted into refined node representations ($H_{\mathcal{G}_{\mathbf{M}}}$, $H_{\mathcal{G}_{\mathbf{D}}}$) encoded by the weight-sharing graph encoders $f_\theta$. Subsequently, they are aggregated into graph representation via the readout layer and finally fed to the weight-sharing graph projection heads $g_\theta$. Thus, the final

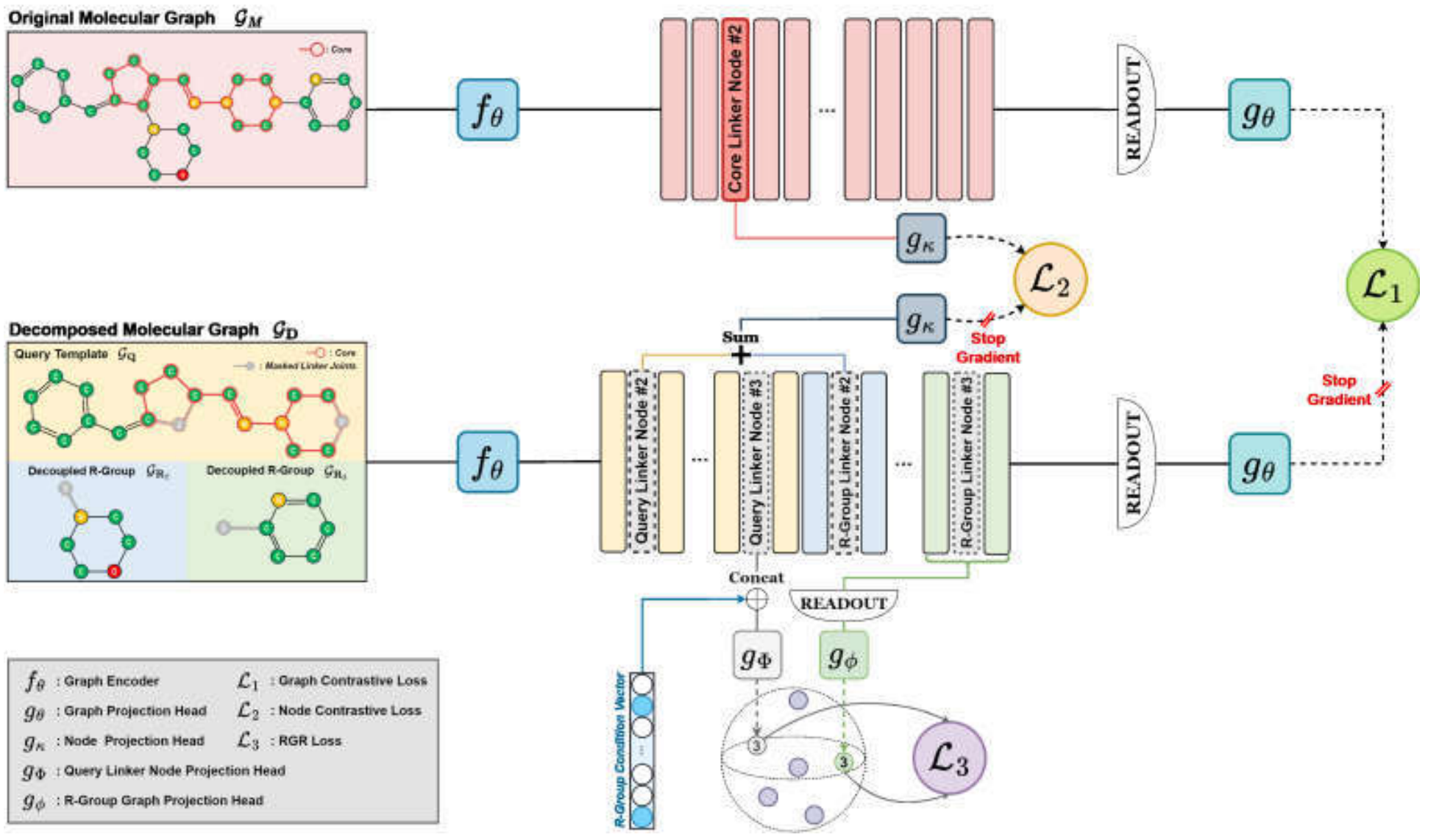

Fig. 2: Overview of *MolPLA* consisting the Masked Graph Contrastive Learning ($\mathcal{L}_1$,$\mathcal{L}_2$) and R-Group Retrieval Framework ($\mathcal{L}_3$). The total loss objective for this pre-training framework is $\mathcal{L} = \mathcal{L}_1 + \mathcal{L}_2 + \mathcal{L}_3$

.

outputs are their corresponding projections in $d$-dimensional molecular graph embedding space ($z_{\mathcal{G}_\mathbf{M}} \in \mathbb{R}^d$, $z_{\mathcal{G}_\mathbf{D}} \in \mathbb{R}^d$).

At the same time, the MGCL framework builds two contrastive node projections which are the core linker node from orignal molecular graph and pair of masked linker nodes from decomposed molecular graph. As shown in Figure 1-(b), the R-groups are decoupled from the putative core leaving both of their linker joints in masked state (i.e. loss of attribute information). *MolPLA* learns how to identify potential decomposable linker joints through alignment of a pair of masked linker nodes with their corresponding node in the original molecular graph.

The $i$th linker node $m_i \in H_{\mathcal{G}_\mathbf{M}}$ selected from the encoded node representations of $\mathcal{G}_\mathbf{M}$ is fed to the weight-sharing node projection heads $g_\kappa$. On the other hand, the corresponding pair of masked linker nodes $q_i, r_i \in H_{\mathcal{G}_\mathbf{D}}$ are selected from the query template $\mathcal{G}_\mathbf{Q}$ and decoupled R-groups $\mathcal{G}_{\mathbf{R}_i}$ with masked linker joints, summed vector-wise ($p_i$) and also fed to the node projection head. As a result, the final outputs are their corresponding projections in $d$-dimensional linker node embedding space ($z_{m_i} \in \mathbb{R}^d$, $z_{p_i} \in \mathbb{R}^d$).

The formulation for this framework that build contrastive graph projections $z_{\mathcal{G}_\mathbf{M}}, z_{\mathcal{G}_\mathbf{D}}$ and node projections $z_{m_i}, z_{p_i}$ is mathematically expressed as follows,

$$H_{\mathcal{G}_\mathbf{M}} = f_\theta(\mathcal{G}_\mathbf{M}) \quad (5)$$

$$H_{\mathcal{G}_\mathbf{D}} = f_\theta(\mathcal{G}_\mathbf{D}) \quad (6)$$

$$z_{\mathcal{G}_\mathbf{M}} = g_\theta(\text{READOUT}(H_\mathbf{M})) \quad (7)$$

$$z_{\mathcal{G}_\mathbf{D}} = \text{STOPGRAD}(g_\theta(\text{READOUT}(H_{\mathcal{G}_\mathbf{D}}))) \quad (8)$$

$$z_{m_i} = g_\kappa(m_i) \quad (9)$$

$$z_{p_i} = \text{STOPGRAD}(g_\kappa(q_i + r_i)) \quad (10)$$

where $H_{\mathcal{G}_\mathbf{M}} \in \mathbb{R}^{n_{\mathcal{G}_\mathbf{M}} \times d}$ and $H_{\mathcal{G}_\mathbf{D}} \in \mathbb{R}^{n_{\mathcal{G}_\mathbf{D}} \times d}$ are the node representations of $\mathcal{G}_\mathbf{M}$ and $\mathcal{G}_\mathbf{D}$ encoded by their respective graph encoders $f_\theta$. $n_{\mathcal{G}_\mathbf{M}}$ and $n_{\mathcal{G}_\mathbf{D}}$ refer to total number of nodes in $\mathcal{G}_\mathbf{M}$ and $\mathcal{G}_\mathbf{D}$.

READOUT and STOPGRAD refer to graph readout function set as global mean pooling and stop-gradient operator. $f_\theta$ is a GNN-based graph encoder with five GIN layers while $g_\theta$ and $g_\kappa$ is a multi-layered perceptron (MLP) with LeakyReLU used as non-linear activation function. The dimensional size $d$ for this framework is fixed to 300. Implementation details for $f_\theta$ are available in our supplementary material **S5**.

### 2.2.2 R-group Retrieval Framework

The RGR Framework builds a co-embedding space of different modalities which are the masked query linker nodes and R-group sub-graphs (Jang *et al.*, 2021). The aim of this framework is to retrieve the most appropriate R-group $\mathcal{G}_\mathbf{R}$ given a linker node of molecular query template graph $\mathcal{G}_\mathbf{Q}$ as input. The retrieved R-groups can be re-attached to their query linker nodes by their respective masked linker joints.

Recall that the weight-sharing graph encoders $f_\theta$ update the node representations of $\mathcal{G}_\mathbf{D}$ where each disjoint sub-graph contains masked linker joints. Among the encoded node representations, the masked query linker node $q_i \in H_{\mathcal{G}_\mathbf{D}}$ is selected from the molecular query template graph. Prior to being propagated to its designated projection head, the query linker node $q_i$ is concatenated with a R-group condition vector $c_{\mathbf{R}_i} \in [0, 1]^{87}$. Motivated from conditional generative modeling, the R-group condition vector guides the RGR framework to retrieve R-groups that fit its encoded criteria that implicate specific functional characteristics (Mirza and Osindero, 2014). At the same time, all nodes are selected from its corresponding R-group $\mathcal{G}_{\mathbf{R}_i}$. They are aggregated by the readout layer (mean pooling) and then fed to the R-group graph projection head $g_\phi$. The final outputs are their respective projections in $d$-dimensional co-embedding space ($z_{q_i} \in \mathbb{R}^d$, $z_{\mathcal{G}_{\mathbf{R}_i}} \in \mathbb{R}^d$).

The formulation for this framework is mathematically expressed as follows,

$$z_{\mathbf{C}_i} = g_\Phi(q_i \oplus c_{\mathbf{R}_i}) \quad (11)$$

$$z_{\mathbf{R}_i} = g_\phi(\text{READOUT}(H_{\mathbf{R}_i})) \quad (12)$$

where both $g_\Phi$ and $g_\phi$ are MLPs with LeakyReLU used as non-linear activation function.

During model inference, the RGR framework is prompted with a query template that consists a decomposed molecular graph containing at least one masked linker joint and user-defined R-group conditions. Each of its masked query linker nodes constrained by its condition vector is projected to the co-embedding space containing over 60,000 R-group embeddings created throughout the pre-training process. Subsequently, the R-groups are retrieved based on their nearest embedding distance with the query linker node projection where the inner product distances are computed by FAISS (Johnson *et al.*, 2019).

#### 2.2.3 Model Optimization

All learning frameworks employ the Dual Information Noise Contrastive Estimation (InfoNCE) objective using in-batch negatives borrowed from Wu *et al.* The InfoNCE loss objective for training *MolPLA* 's both frameworks given two different projections $x$, $y$ as input is mathematically expressed as follows,

$$\mathcal{L} = -\frac{1}{2\mathcal{B}}\sum_{i=1}^{\mathcal{B}}[\log\frac{\exp(\text{sim}(x_i,y_i)/\tau)}{\sum_{j=1}^{\mathcal{B}}\exp(\text{sim}(x_i,y_j)/\tau)} + \log\frac{\exp(\text{sim}(y_i,x_i)/\tau)}{\sum_{j=1}^{\mathcal{B}}\exp(\text{sim}(y_i,x_j)/\tau)}] \quad (13)$$

where $\mathcal{B}$ is number of projection pairs within a batch and $\tau$ is temperature scalar for adjusting the uniformity of embedding space. The similarity score function sim$(\cdot,\cdot)$ in this loss objective is cosine similarity between two vector projections.

For the MGCL framework's graph contrastive loss $\mathcal{L}_1$, the two input projections for the InfoNCE term are those of original and decomposed molecular graph ($x{=}z_{\mathcal{G}_\mathbf{M}}$, $y{=}z_{\mathcal{G}_\mathbf{D}}$). $\mathcal{B}_1$ is number of data instances per batch while $\tau_1$ is set to 0.01. For MGCL framework's node contrastive loss $\mathcal{L}_2$, the two input projections are those of original linker node and masked linker node pair($x{=}z_m$, $y{=}z_p$). $\mathcal{B}_2$ is total number of linker nodes in original molecule per batch while $\tau_2$ is set to 0.05. For the RGR framework $\mathcal{L}_3$, the two input projections are those of query linker node and its previously linked R-group sub-graph ($x{=}z_\mathbf{C}$, $y{=}z_\mathbf{R}$). $\mathcal{B}_3$ is total number of query linker and R-group pairs per batch while $\tau_3$ is set to 0.01. Since each data instance can have one or more of decomposed R-groups due to the proposed molecular decomposition method, $\mathcal{B}_1 \geq \mathcal{B}_2$ and $\mathcal{B}_1 \geq \mathcal{B}_3$.

The total loss objective for *MolPLA* is sum of the three InfoNCE loss terms which is mathematically expressed as follows,

$$\mathcal{L} = \mathcal{L}_1 + \mathcal{L}_2 + \mathcal{L}_3 \quad (14)$$

While each of the loss terms have its own optimization trajectory with respect to different pairs of projections, the parameters of the weight-sharing graph encoder $f_\theta$ are updated by all three types of back-propagated gradients. We expect this multi-faceted optimization approach facilitate the creation of more complex and intriguing graph representations.

## 3 Results

### 3.1 Experimental Settings

We conducted experiments to evaluate *MolPLA* 's performance on molecule property prediction and R-group retrieval task compared with its baseline and ablated models. We pre-trained all model variants on the same GEOM molecules. For *MolPLA* , we sampled and reserved 10% of our pre-processed GEOM dataset as hold-out test set. The purpose of the test set is to evaluate *MolPLA* on the R-Group retrieval task since the baseline pre-training methods do not feature the R-Group Retrieval framework.

Eleven baseline models were used in the molecule property prediction task for comparative evaluation with our proposed model (AttrMask (Hu *et al.*, 2019), EdgePred (Hamilton *et al.*, 2017), InfoGraph (Sun *et al.*, 2019), GraphCL (You *et al.*, 2020), GPT-GNN (Hu *et al.*, 2020), GROVER (Rong *et al.*, 2020), MGSSL (Zhang *et al.*, 2021), MolCLR (Wang *et al.*, 2022b), iMolCLR (Wang *et al.*, 2022a) and FREL (Wu *et al.*, 2023). Having imported their model hyperparameters and additional data augmentation methods from their original code repositories, we pre-trained these models on all existing molecules in the GEOM dataset from scratch and saved their checkpoints.

The ablated versions of *MolPLA* are **MolPLA**$_{w/o\ RGR}$ and **MolPLA**$_{Murcko}$. For **MolPLA**$_{Murcko}$, the putative cores identified from Naveja *et al.*'s framework are replaced with Murcko scaffolds during the graph decomposition process.

We pre-trained *MolPLA* and its ablations to a maximum of 100 epochs with early stopping whose criteria is based on the per-epoch validation loss. With batch size set to 512, we used Adam optimizer with learning rate set to 0.001. During fine-tuning on all molecular property prediction benchmark datasets, we trained them to a maximum of 100 epochs with early stopping criteria set to either per-epoch validation AUROC or RMSE. With batch size and dropout rate set to 32 and 0.3, we used Adam optimizer with learning rate set to 0.0001. This hyperparameter setting was identically used in all benchmark tasks.

### 3.2 Molecule Property Prediction Task

Nine molecule property datasets were used in benchmark evaluation where six of them are classification (**Bace**, **BBBP**, **ClinTox**, **Sider**, **Tox21**, **Toxcast**) and three of them are regression tasks (**ESOL**, **FreeSolv**, **Lipophilicity**). The evaluation metrics are Area Under ROC (AUROC) and Root Mean-Squared Error (RMSE) respectively.

We loaded the pre-trained weights of *MolPLA* 's Graph Encoder $f_\theta$ and added a simple linear layer denoted as Task Prediction Head where its output dimension depends on the number of tasks in a benchmark dataset. We then fine-tuned it during the training phase on each benchmark dataset. We performed dataset partition with a train-validation-test ratio of 8:1:1 based on the scaffold split method employed by previous works. We repeated this experiment five times with different random seeds and calculated the mean and standard deviation of evaluation scores. The same procedure was done to its baseline and ablated models as well.

According to Table 2, **MolPLA** achieved the second-best and best performance in terms of AUROC (0.7204) and RMSE (1.6021) averaged across all benchmark datasets, while besting other models in **BBBP** and **FreeSolv** dataset. In fact, the ablated version of *MolPLA* without the R-group retrieval functionality (**MolPLA**$_{w/o\ RGR}$) showed superior performance in classification tasks (0.7312), achieving state-of-art performance in the **ClinTox** (0.8973) and **Tox21** (0.7536) dataset. Overall, the proposed MGCL framework in *MolPLA* demonstrated its effectiveness of graph representation learning based on molecular cores, R-groups and linker joints by helping the model generalize in molecule property prediction tasks.

| Model Description | | Classification (AUROC↑) | | | | | | | Regression (RMSE↓) | | | |
|---|---|---|---|---|---|---|---|---|---|---|---|---|
| | | Bace | BBBP | ClinTox | Sider | Tox21 | ToxCast | Average | ESOL | FreeSolv | Lipophilicity | Average |
| MolPLA and its Ablations | MolPLA | 0.7785 (0.0244) | **0.7054 (0.0166)** | 0.8422 (0.0327) | 0.6080 (0.0081) | 0.7524 (0.0048) | 0.6356 (0.0042) | 0.7204 | 1.3837 (0.0441) | **2.6061 (0.1681)** | 0.8165 (0.0087) | **1.6021** |
| | MolPLA$_{w/o\ RGR}$ | 0.8046 (0.0166) | 0.7018 (0.0182) | **0.8973 (0.0215)** | 0.5941 (0.0092) | **0.7536 (0.0056)** | 0.6358 (0.0047) | **0.7312** | 1.3372 (0.0627) | 3.1832 (0.1304) | 0.7953 (0.0086) | 1.7719 |
| | MolPLA$_{Murcko}$ | 0.7864 (0.0207) | 0.6610 (0.0275) | 0.8574 (0.0355) | **0.6085 (0.0054)** | 0.7477 (0.0046) | 0.6343 (0.0036) | 0.7159 | 1.3923 (0.0871) | 2.6601 (0.0871) | 0.8243 (0.0149) | 1.6256 |
| General Graph Pre-training Baselines | AttrMask | 0.7840 (0.0174) | 0.6619 (0.0259) | 0.7629 (0.0170) | 0.5365 (0.0134) | 0.7072 (0.0104) | 0.5566 (0.0046) | 0.6682 | **1.3012 (0.0410)** | 3.8309 (0.1887) | 0.8092 (0.0155) | 1.9804 |
| | EdgePred | 0.7371 (0.0160) | 0.6919 (0.0110) | 0.6205 (0.0109) | 0.6079 (0.0040) | 0.6984 (0.0044) | 0.5757 (0.0037) | 0.6553 | 1.4116 (0.0301) | 3.9738 (0.1710) | 0.8884 (0.0109) | 2.0913 |
| | InfoGraph | 0.7356 (0.0076) | 0.6546 (0.0225) | 0.8354 (0.0348) | 0.5934 (0.0103) | 0.7022 (0.0081) | 0.5523 (0.0079) | 0.6789 | 1.4283 (0.0651) | 5.4324 (0.1187) | 0.8030 (0.0208) | 2.5546 |
| | ContextPred | 0.7611 (0.0119) | 0.6511 (0.0114) | 0.6015 (0.0165) | 0.6054 (0.0059) | 0.7086 (0.0016) | 0.5736 (0.0017) | 0.6502 | 1.5568 (0.0271) | 4.1888 (0.1501) | 0.8242 (0.0074) | 2.1899 |
| | GraphCL | 0.7170 (0.0090) | 0.6714 (0.0027) | 0.7267 (0.0375) | 0.5968 (0.0181) | 0.7212 (0.0069) | 0.5688 (0.0040) | 0.6670 | 1.3868 (0.0683) | 3.4046 (0.1716) | 0.7859 (0.0142) | 1.8591 |
| | GPT-GNN | 0.7745 (0.0216) | 0.6939 (0.0077) | 0.7467 (0.0269) | 0.5683 (0.0198) | 0.7314 (0.0100) | 0.5642 (0.0076) | 0.6798 | 1.3761 (0.0384) | 3.6846 (0.4096) | 0.7757 (0.0170) | 1.9455 |
| Molecular Graph Pre-training Baselines | GROVER | 0.7439 (0.0172) | 0.6669 (0.0178) | 0.8055 (0.0276) | 0.5920 (0.0092) | 0.7094 (0.0077) | 0.5704 (0.0047) | 0.6814 | 1.5355 (0.0683) | 4.0187 (0.1820) | 0.8016 (0.0343) | 2.1186 |
| | MGSSL | 0.7863 (0.0069) | 0.6647 (0.0107) | 0.6913 (0.1198) | 0.5983 (0.0076) | 0.7481 (0.0068) | 0.6309 (0.0031) | 0.6866 | 1.3596 (0.0457) | 2.7975 (0.1541) | **0.7649 (0.0171)** | 1.6407 |
| | MolCLR | 0.7529 (0.0174) | 0.6459 (0.0204) | 0.5876 (0.0189) | 0.5857 (0.0050) | 0.7300 (0.0069) | 0.5600 (0.0062) | 0.6437 | 1.3794 (0.0425) | 3.2484 (0.1635) | 0.7729 (0.0127) | 1.8002 |
| | iMolCLR | 0.7566 (0.0090) | 0.6691 (0.0230) | 0.6478 (0.0193) | 0.5866 (0.0073) | 0.7157 (0.0099) | 0.5539 (0.0091) | 0.6550 | 1.4043 (0.0285) | 3.3375 (0.2204) | 0.7709 (0.0064) | 1.8376 |
| | FREL | **0.8107 (0.0294)** | 0.6624 (0.0230) | 0.6806 (0.0384) | 0.5966 (0.0057) | 0.7380 (0.0046) | **0.6384 (0.0062)** | 0.6878 | 1.7204 (0.0471) | 4.7942 (0.6774) | 1.0022 (0.0155) | 2.5056 |

Table 2. Experimental results on nine benchmark datasets for molecular property prediction task. The best and second-best scores for each dataset are in bold-faced and underlined.

| Model | GEOM (Pre-training Dataset) | | | | DrugBank (External Dataset) | | | |
|---|---|---|---|---|---|---|---|---|
| | MRR | R@10 | R@50 | R@100 | MRR | R@10 | R@50 | R@100 |
| **MolPLA** | **0.2616** | **0.4839** | **0.7711** | **0.8702** | 0.2213 | **0.4309** | **0.6643** | **0.7378** |
| **Popularity** | 0.0454 | 0.0547 | 0.1669 | 0.3814 | **0.2448** | 0.2696 | 0.3311 | 0.4381 |
| **Random** | 0.0014 | 0.0001 | 0.0003 | 0.0015 | 0.0006 | 0.0001 | 0.0002 | 0.0014 |
| **Cond. None** | 0.0056 | 0.0105 | 0.0207 | 0.0272 | 0.0032 | 0.0056 | 0.0125 | 0.0158 |
| **Cond. All** | <0.0001 | <0.0001 | 0.0001 | 0.0002 | <0.0001 | 0.0001 | 0.0004 | 0.0005 |

Table 3. Experimental results on the GEOM and DrugBank dataset for R-group retrieval task.

## 3.3 R-Group Retrieval Task

We evaluated *MolPLA* 's performance on R-group retrieval task by running model inference on the reserved data instances of our GEOM-based pre-training dataset. The main objective of this task is to assess whether *MolPLA* can retrieve appropriate R-groups given a query template as input. We first constructed a library containing 61,279 R-groups by running inference on *MolPLA* 's pre-trained graph encoder $f_\theta$ and graph projection head $g_\theta$ with their corresponding graph inputs with masked linker joints. For each test data instance which contains a molecular query template graph with masked linker nodes and its corresponding decoupled R-groups as ground truths, we first encoded the query template into its node representations via the same pre-trained graph encoder. We projected the query linker nodes into the pre-trained co-embedding space and retrieved the top 1000 nearest R-groups using FAISS with vector-wise inner product distance.

As each masked linker node was originally bonded with its actual R-group prior to decomposition, the pre-trained model is expected to retrieve that R-group along with other possible candidates. We used mean reciprocal rank (MRR) and recall at K (R@5, 50, 100) to evaluate *MolPLA* 's R-group retrieval task. We further evaluated *MolPLA* 's R-group retrieval task on external dataset DrugBank containing 8,288 drug compounds. The same pre-processing method and R-group library was used for this dataset to assess *MolPLA* 's generalizability in unseen molecules.

We included two naive guessing baselines which are **Random Choice** and **Popularity Choice**. The former refers to randomly selecting a fixed number of R-groups from the library while the latter refers to giving each input the same list of top 1000 frequent R-groups (`*∼O`, `*∼c1ccccc1`). We additionally included ablations where the R-group condition vector is either filled with only 0-bits (**Cond. None**) or 1-bits (**Cond. All**). According to the experimental results on both datasets shown in Table 3, *MolPLA* exhibited its robustness in retrieving R-groups when given various query core templates.

## 3.4 Qualitative Analysis

We investigated *MolPLA* 's understanding in the underlying structural concepts of chemical compounds by analyzing its inherently created node representations and R-group library. In addition, we deployed *MolPLA* in a lead optimization scenario where it generates novel molecules with improved properties and docking results. The example reference molecules used in the qualitative analysis are Streptozocin and Capmatinib which were selected from the external DrugBank dataset. Additional qualitative analysis results on other DrugBank compounds, Riluzole and Lasmiditan are available in the supplementary material **S7**, **S8**, **S9**.

### 3.4.1 Visualization of Node Representations

The central motivation of *MolPLA* 's MGCL framework is pushing its molecular representation learning towards decomposition of core and R-group substructures through identification of linker joints. For each reference molecule, we used Principal Component Analysis on its node representations extracted from *MolPLA* 's graph encoder $f_\theta$ and used the first dimensionality reduced values for graph node coloring analysis.

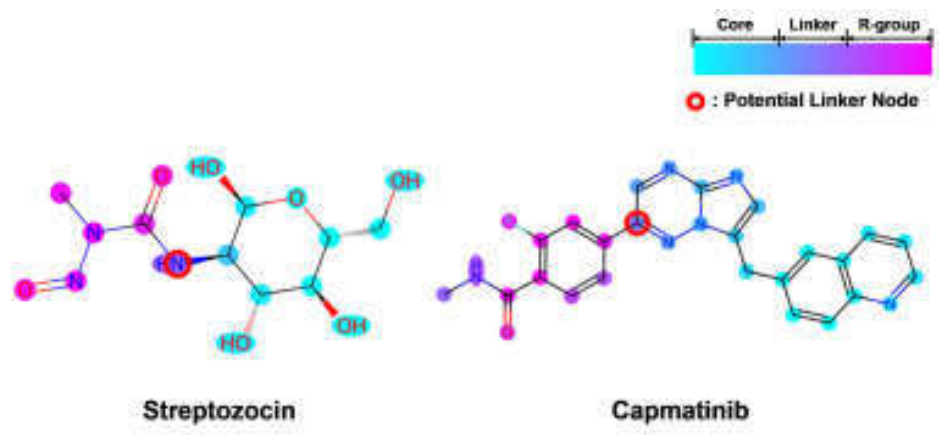

Fig. 3: Visualization results of node representations generated by *MolPLA* for the two reference molecules Streptozocin and Capmatinib.

Figure 3 shows the node coloring results on the two reference molecules. The node colorings seemed to exhibit regional consistency where there are distinct coloring differences between two sub-structures (molecular core and R-group) of each molecule. We speculate that the node representations from *MolPLA* 's pre-trained graph encoder draw distinction between core and R-group sub-structures which naturally implicates potential linker nodes used for their decomposition.

### 3.4.2 Deployment of *MolPLA* in Lead Optimization

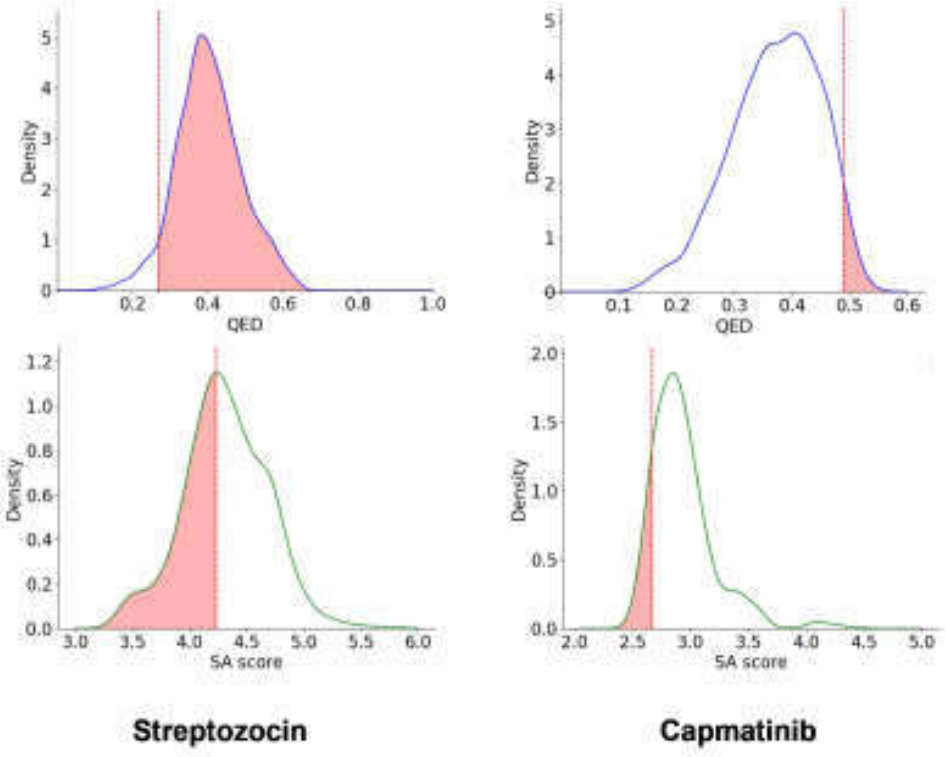

Fig. 4: Distribution of QED and SAscore generated by *MolPLA* for the two reference molecules Streptozocin and Capmatinib. The red dashed line represents the scores of the original molecules, while the red region indicates a subset of values from the optimized molecules.

Lead optimization in drug discovery involves replacement of R-groups while pivoting the reference molecule's core structure and other R-groups. We employed *MolPLA* 's RGR framework in optimizing the above-mentioned reference molecules Streptozocin and Capmatinib. Based on the node coloring results in Figure 3, we first selected the linker nodes used for applying molecular graph decomposition to the reference molecules and obtained each of its molecular query template graph and decoupled R-group sub-graph with masked linker joints. The designated query template graph with its R-group to be substituted is `*∼[C@@H]1[C@@H](O)[C@H](O)[C@@H](CO)O[C@@H]1O`, `*∼C(=O)N(C)N=O` for Streptozocin and `*∼c1:cnc2ncc(Cc3ccc4ncccc4c3)n2n:1`, `*∼c1ccc(C(=O)NC)c(F)c1` for Capmatinib.

We then used *MolPLA* 's RGR framework by projecting the query template's linker node concatenated with its condition vector to co-embedding space containing the pre-trained R-group graph representations. The top 1000 R-groups based on each of its calculated vector distance with the query linker node were retrieved and re-attached to the query template based on their matching masked linker joints. During the re-attachment process, the masked linker joints are filled with proper node and edge attributes to ensure the validness of the generated molecule. Detailed explanation of this procedure is available in the supplementary material **S6**. The proportion of valid molecules with improved quantitative estimate of drug-likeness (QED) and synthetic accessibility score (SAscore) are 97% for Streptozocin and 9% for Capmatinib. Figure 4 shows distribution of calculated scores of validly generated molecules.

In addition to improvement in QED and SA scores, our findings also demonstrate improvement in binding properties with target protein, which interacts with the reference molecule, and the molecules we generated. We first obtained their corresponding binding complexes from the RCSB database which are `2w4x` and `5ya5` for Streptozocin and Capmatinib respectively (Burley *et al.*, 2023). The binding target in `2w4x` is BtGH84, a homolog of human O-GlcNAcase from *Bacteroides thetaiotaomicron* (Porter *et al.*, 2018). The target in `5ya5` is c-Met kinase, one of the main targets for developing antitumor agents (Yuan *et al.*, 2018).

We performed docking simulations using Maestro from the Schrödinger package (Maestro, 2020) on each of its protein pocket with the generated molecules to obtain their docking scores and poses. Figure 5 presents exemplary docking results compared to their reference molecules. Judging from improvement in their drug-likeness, docking scores and semblance of their poses, *MolPLA* demonstrated its potential of contributing to lead optimization scenarios in drug discovery.

### 3.4.3 Cross-Reference R-Group Retrieval Results

The RGR framework in *MolPLA* creates a query linker node projection based on integration of external R-group conditions and query template graph context previously encoded by $f_\theta$. We performed cross-reference analysis to assess *MolPLA* 's query node projection head's $g_\Phi$ sensitivity against different R-group conditions and query template graphs. From perspective of recommender systems, retrieval results that are exclusively sensitive to either the R-group conditions or the query template graph context, but not both, are considered suboptimal.

As shown in Table 4, the R-group retrieval results for Streptozocin's query template given R-group condition vector from Capmatinib differ from those obtained using its originally decoupled R-group's condition vector. This observation is reciprocally true for Capmatinib. More cross-reference analyses using two additional reference molecules Riluzole and Lasmiditan are available in the supplementary material **S10**. Our investigation in these results reveal that *MolPLA* 's RGR framework is indeed senstive to variations in both R-group conditions and the query template graph context, which underlines *MolPLA* 's potential in generating a wider range of novel compounds.

## 4 Discussion

### 4.1 Impact of Proposed Masked Graph Contrastive Learning Framework

Having incorporated the concepts of molecular cores and R-groups, we employed Naveja *et al.*'s putative core framework for devising a novel molecular graph decomposition method. Since the proposed decomposition method generates multiple results, several advantages related to representation learning were confirmed in our studies including data augmentation effects and generalizability in novel molecules. Furthermore, utilizing masked linker joints enabled *MolPLA* to effectively discern potential cores, R-groups, and their corresponding decomposable regions as linker joints. The MGCL framework played a key role in shaping *MolPLA* 's molecular comprehension, particularly in identifying these key structural components.

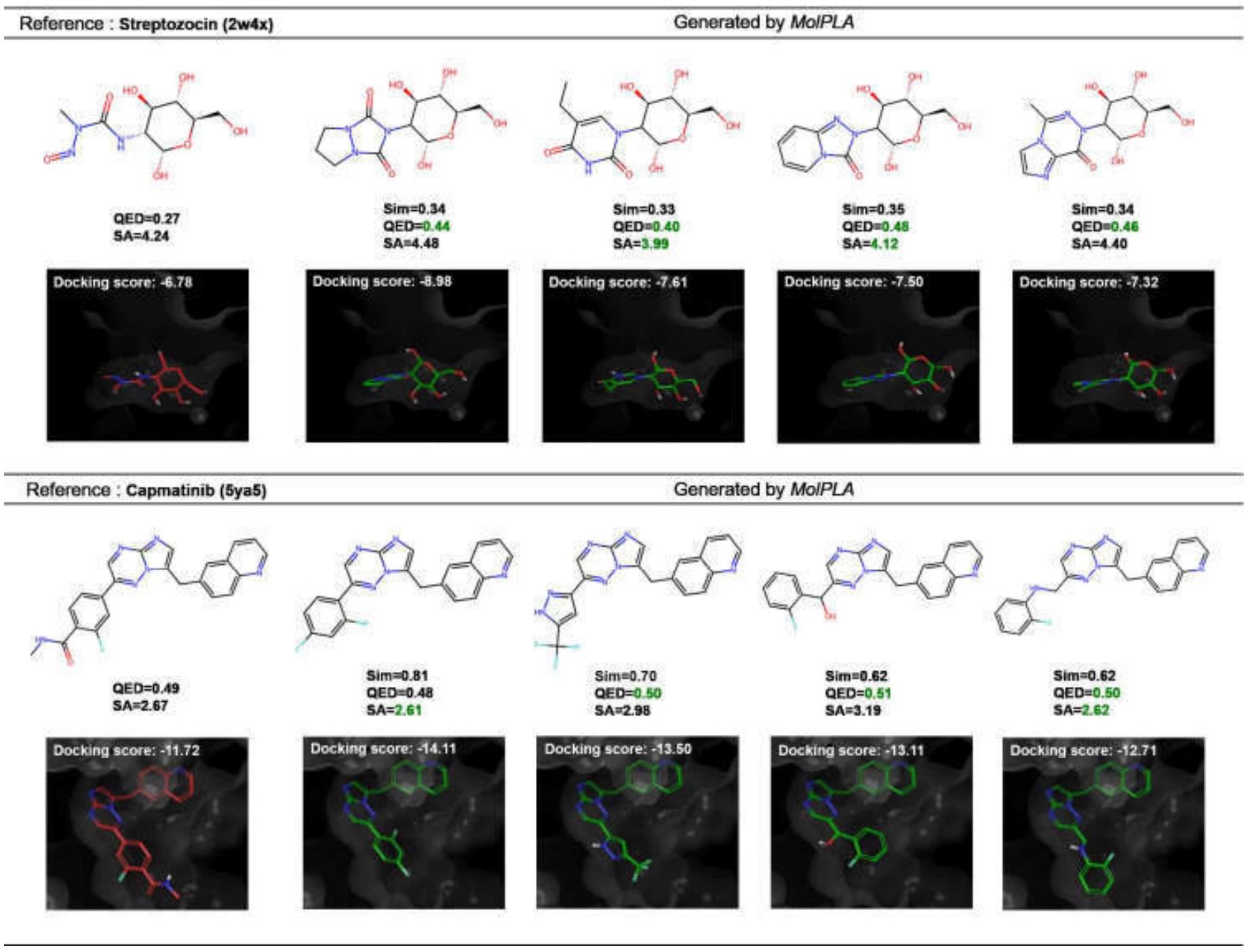

Fig. 5: List of generated molecules for each reference molecule in lead optimization scenario. The generated molecules were selected based on their calculated drug-likeness scores which are QED, SA Score and Docking Score

| Query Template | R-group Substitution | Top 5 Retrieved R-groups (SMILES) |
|---|---|---|
| **Streptozocin** | **Streptozocin** | *∼n1oc(=O)[nH]c1=O, *∼n1ccc(=O)[nH]c1=O, *∼n1ncc(=O)[nH]c1=O<br>*∼n1c(=O)[nH]c(=O)n(C)c1=O, *∼OCc1ccccc1[N+](=O)[O-] |
| **Streptozocin** | **Capmatinib** | *∼OCc1ccccc1F, *∼OCc1ccc(F)cc1, *∼OCc1c(F)cccc1F<br>*∼NOC(=O)c1ccccc1F, *∼OCc1cc(F)ccc1-c1ccc(C)cc1 |
| **Capmatinib** | **Capmatinib** | *∼c1ccc(F)c(C(F)(F)F)c1, *∼c1ccc(C(F)(F)F)c(F)c1, *∼c1cccc(C(F)(F)F)c1<br>*∼c1ccc(F)c2ccccc21, *∼c1cc(C(F)(F)F)cc(C(F)(F)F)c1 |
| **Capmatinib** | **Streptozocin** | *∼n1nnc(C(=O)/C=C/N(C)C)c1C, *∼n1nc(C)c(C#N)c1N, *∼n1nc(C)c(CN)c1C<br>*∼n1nc(C)c(NC(=O)c2ccnn2C)c1C, *∼n1nc(C)c(NC(=O)c2ccn(CC)n2)c1C |

Table 4. Cross reference analysis results based on switching one's decoupled R-group's condition vector with the other when retrieving R-group suggestions through MolPLA 's RGR framework.

## 4.2 Impact of Proposed R-Group Retrieval Framework in Lead Optimization

The proposed RGR framework provides several advantages that can potentially benefit lead optimization scenarios. *MolPLA* is capable of making rationale R-group suggestions based on its joint understanding of the molecular query template's context and R-group conditions that can be modified according to the user's needs. Docking simulations show promising results of the RGR framework's utility in retrieving plausible R-group substitutions involved in our lead optimization scenarios.

## 4.3 Limitations and Future Work

Quantitative results on molecular property prediction tasks do not fully demonstrate the advantages of simultaneously optimizing both of *MolPLA* 's frameworks. This could be due to lack of effort for resolving the adversarial optimization trajectories incurred by three different loss objectives during the pre-training phase. We plan to carefully design a optimal method to ensure *MolPLA* gains synergistic benefits from both MGCL and RGR frameworks.

Despite our advantages of employing our molecular graph decomposition method for data augmentation during construction of the pre-training dataset, the 300K compounds from GEOM are biased and insufficient compared to other large-scale pre-training practices in other domains. We plan to pre-train *MolPLA* on a larger database such as ZINC15 or PubChem to amplify its structural understanding in vast compounds.

Another limitation related to the RGR framework is its inability to take protein pocket structures into account. In drug generation scenarios, considering not only resemblance to reference molecule but also its corresponding target protein is crucial. As part of future work, we plan to incorporate drug-target binding complexes and formulate R-group retrieval conditioned by protein pockets.

## 5 Conclusion

In this work, we introduce *MolPLA* , a novel pre-training framework that promotes deep understanding in global and local molecular structures which are cores, R-groups and decomposable linker joints. *MolPLA* consists of two frameworks which are the Masked Graph Contrastive Learning and R-Group Retrieval frameworks. Quantitative results on molecular property prediction tasks and qualitative analysis including docking results demonstrate the potential benefits of utilizing *MolPLA* in drug discovery.

## Funding

This work was supported in part by the National Research Foundation of Korea [NRF-2023R1A2C3004176], the Ministry of Health & Welfare, Republic of Korea [HR20C0021(3)], the Ministry of Science and ICT (MSIT) [RS-2023-00262002], and the ICT Creative Consilience program through the Institute of Information & Communications Technology Planning & Evaluation(IITP) grant funded by the MSIT [IITP-2024-2020-0-01819].

# Supplementary Material for **MolPLA:** *A Molecular Pre-training Framework for Learning Cores, R-Groups and their Linker Joints*

## **S1:** Detailed statistics of the MolPLA pre-training dataset

**Original GEOM: 304,466 molecules**

### 1. After core extraction

(GEOM mol, core): 1,231,464 pairs

**Molecule Core counts**

: Distribution and percentile of core counts extracted from GEOM molecules

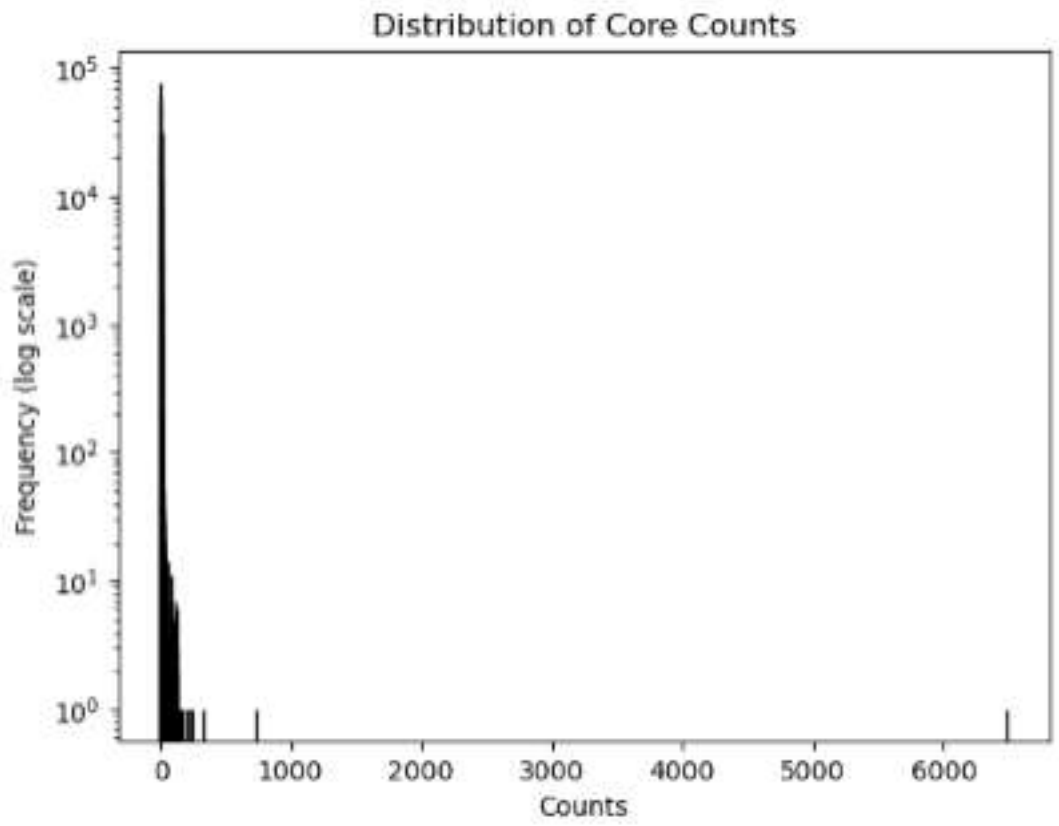

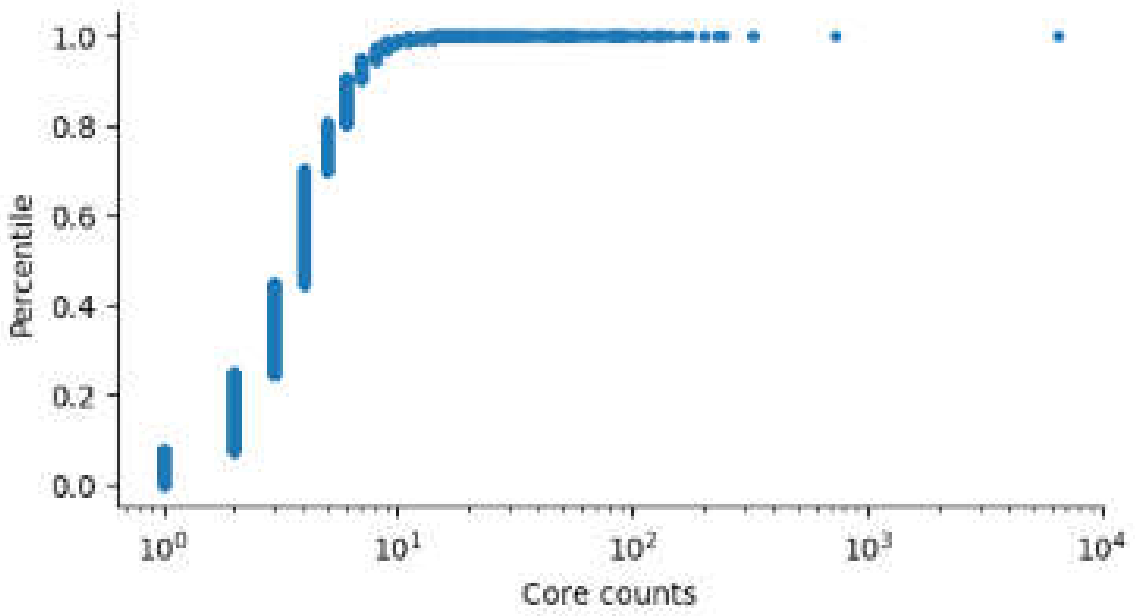

- Max: 6486, Min: 1, Median: 4, Mean: 4.05, Mode: 4

### 2. After applying core number threshold

(GEOM mol, core): 1,196,157 pairs

Apply a core number threshold of 10 for each molecule, randomly selecting cores for those with more than the threshold.

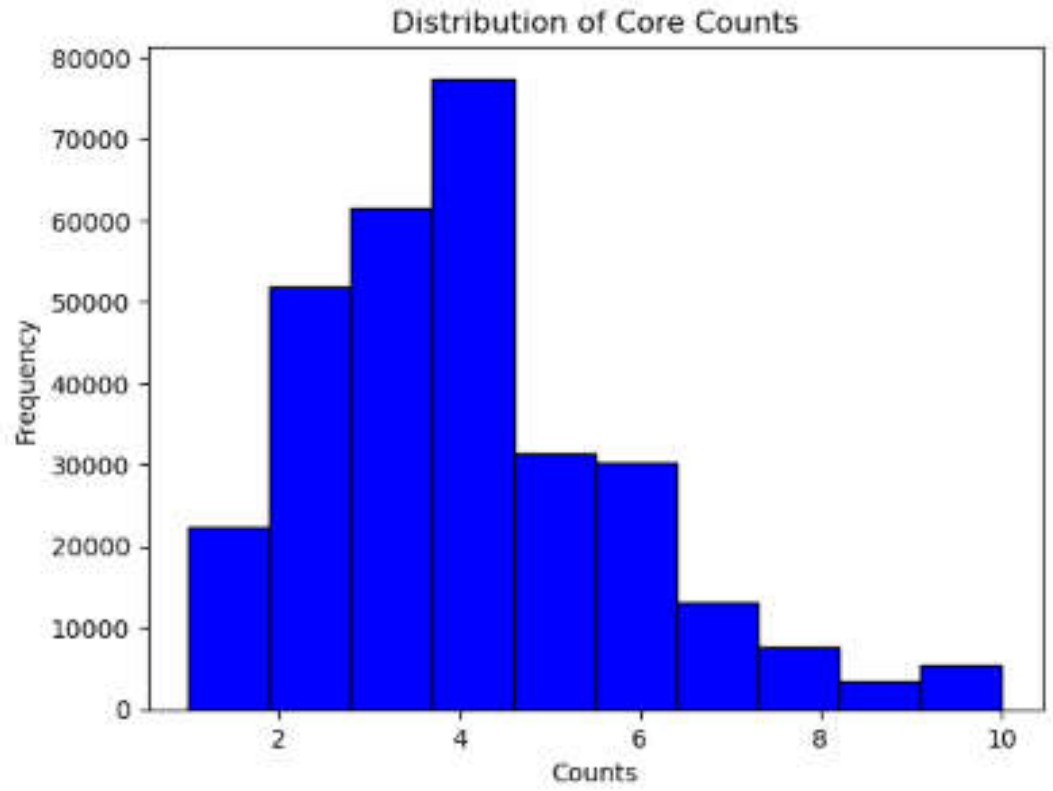

- Max: 10, Min: 1, Median: 4, Mean: 3.85, Mode: 4

# 3. After R-Group attachment

(GEOM mol, core, attached R-Groups): 1,259,946 pairs

**Top 80 Putative Cores histogram**

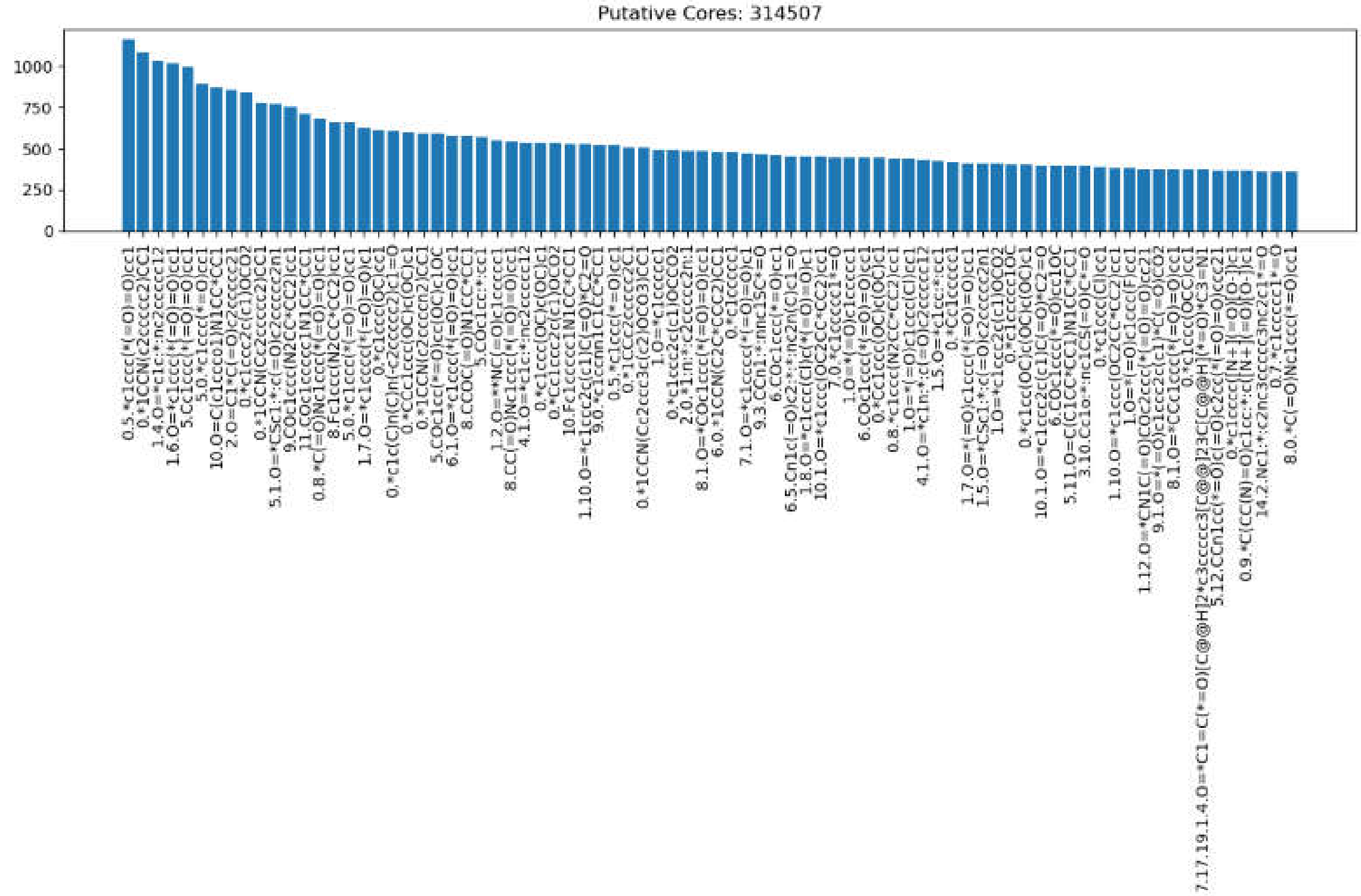

**Top 20 Putative Cores image**

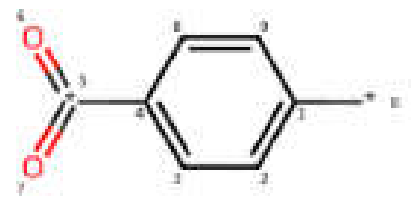

0.5.*c1ccc(*(=O)=O)cc1: 1164

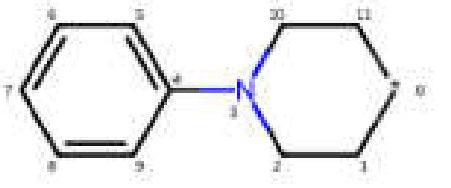

0.*1CCN(c2ccccc2)CC1: 1088

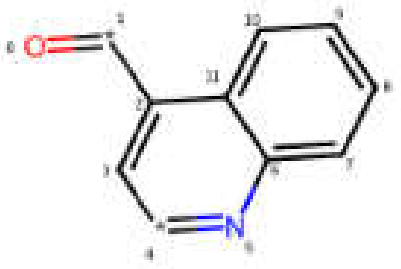

1.4.O=*c1c:*:nc2ccccc12: 1032

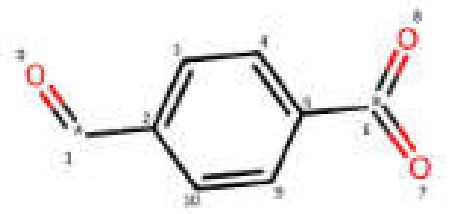

1.6.O=*c1ccc(*(=O)=O)cc1: 1017

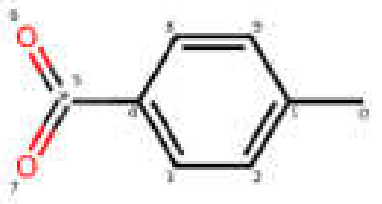

5.Cc1ccc(*(=O)=O)cc1: 994

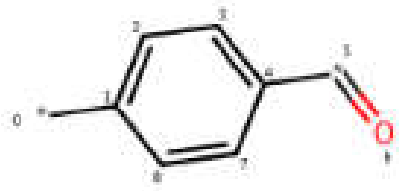

5.0.*c1ccc(*=O)cc1: 888

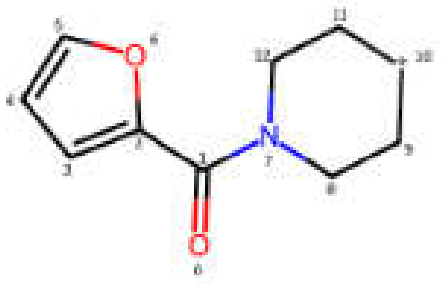

10.O=C(c1ccco1)N1CC*CC1: 868

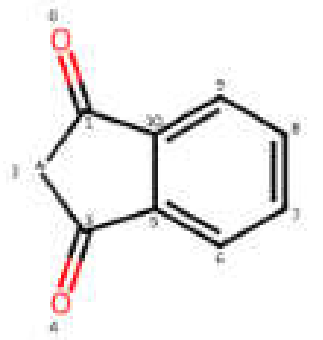

2.O=C1*C(=O)c2ccccc21: 857

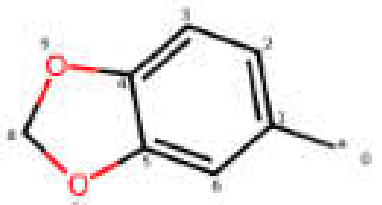

0.*c1ccc2c(c1)OCO2: 840

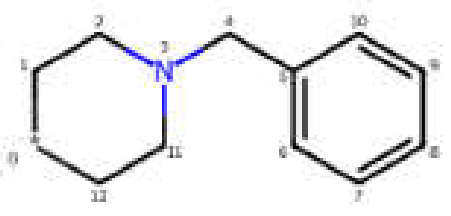

0.*1CCN(Cc2ccccc2)CC1: 779

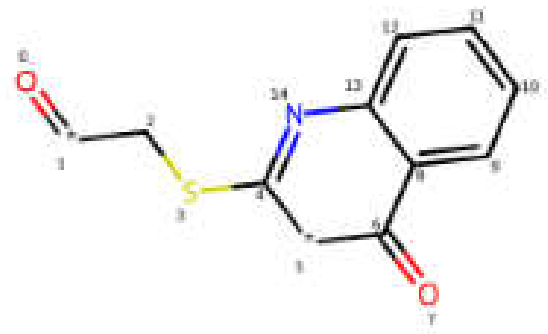

5.1.O=*CSc1:*:c(=O)c2ccccc2n1: 771

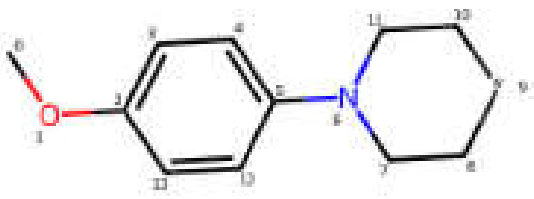

9.COc1ccc(N2CC*CC2)cc1: 753

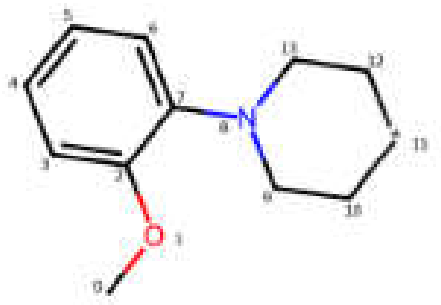

11.COc1ccccc1N1CC*CC1: 710

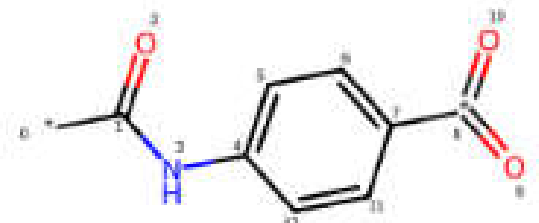

0.8.*C(=O)Nc1ccc(*(=O)=O)cc1: 681

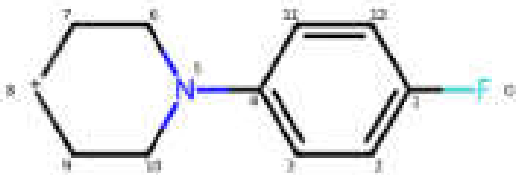

8.Fc1ccc(N2CC*CC2)cc1: 661

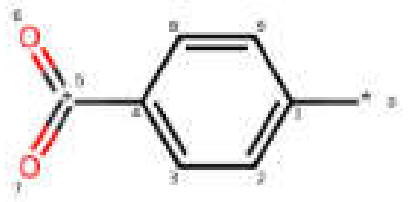

5.0.*c1ccc(*(=O)=O)cc1: 657

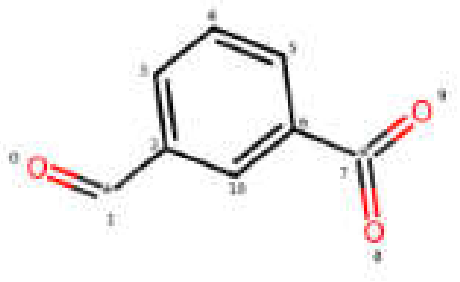

1.7.O=*c1cccc(*(=O)=O)c1: 624

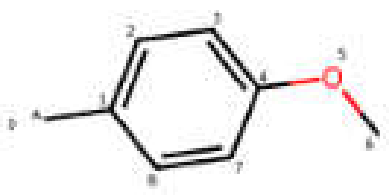

0.*c1ccc(OC)cc1: 610

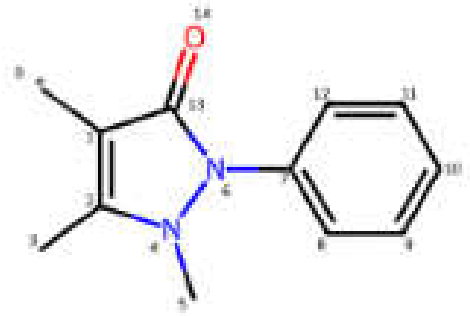

0.*c1c(C)n(C)n(-c2ccccc2)c1=O: 605

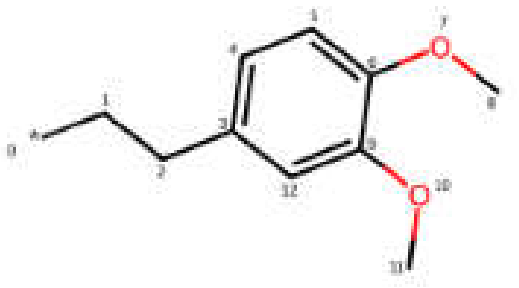

0.*CCc1ccc(OC)c(OC)c1: 598

**Top 80 R-Groups histogram**

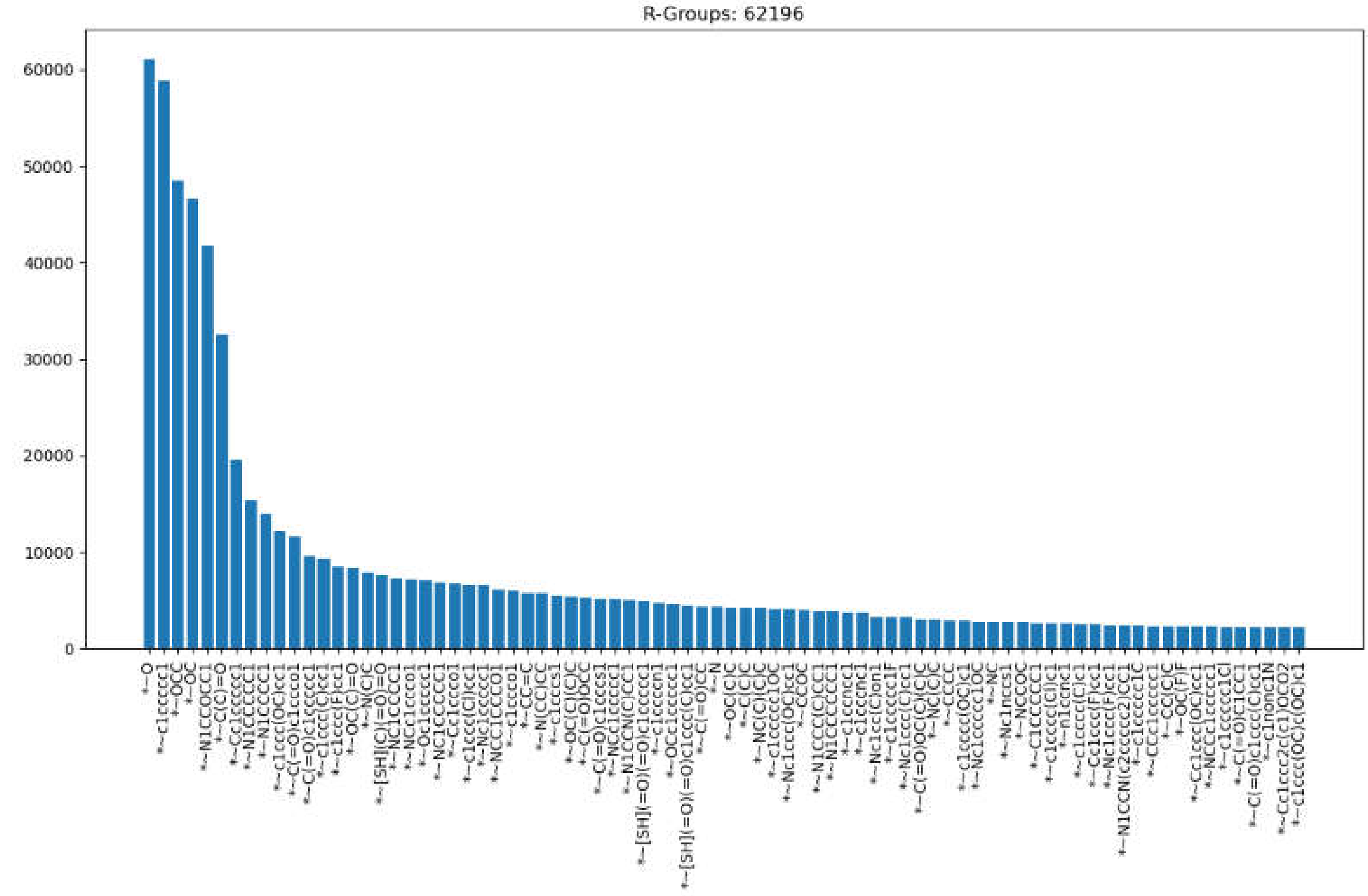

**Top 20 R-groups image**

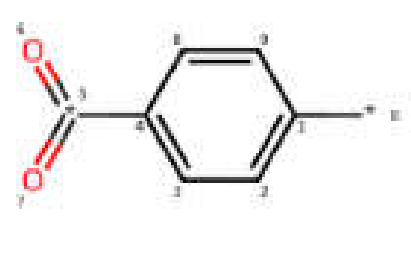

0.5.*c1ccc(*(=O)=O)cc1: 1164

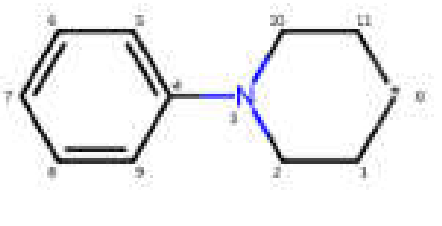

0.*1CCN(c2ccccc2)CC1: 1088

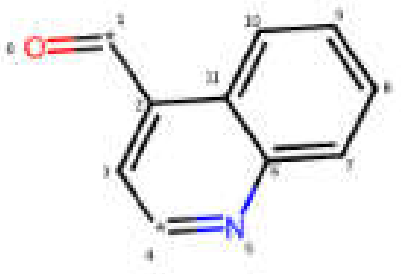

1.4.O=*c1c:*:nc2ccccc12: 1032

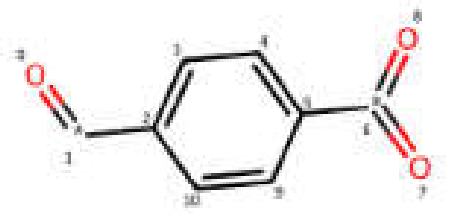

1.6.O=*c1ccc(*(=O)=O)cc1: 1017

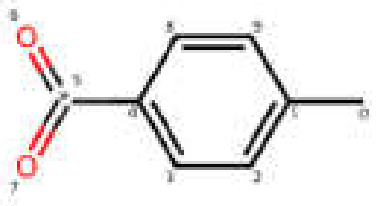

5.Cc1ccc(*(=O)=O)cc1: 994

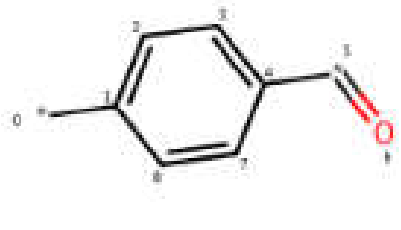

5.0.*c1ccc(*=O)cc1: 888

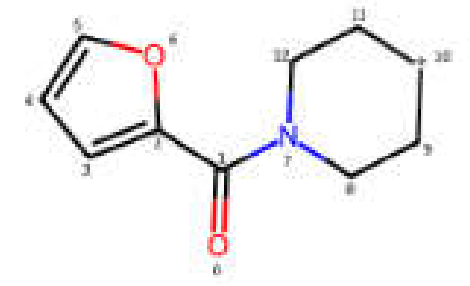

10.O=C(c1ccco1)N1CC*CC1: 868

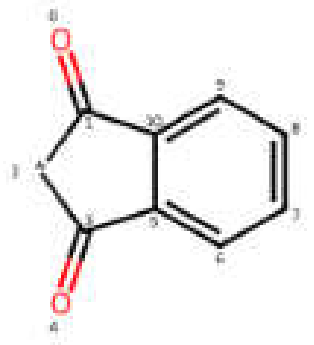

2.O=C1*C(=O)c2ccccc21: 857

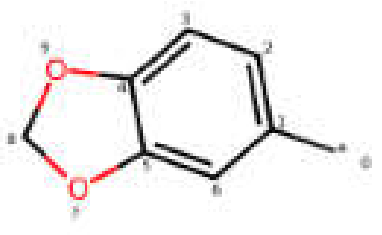

0.*c1ccc2c(c1)OCO2: 840

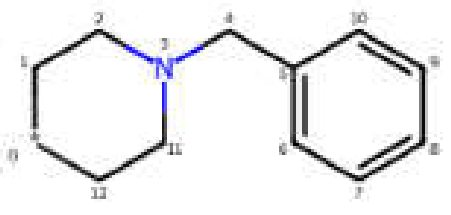

0.*1CCN(Cc2ccccc2)CC1: 779

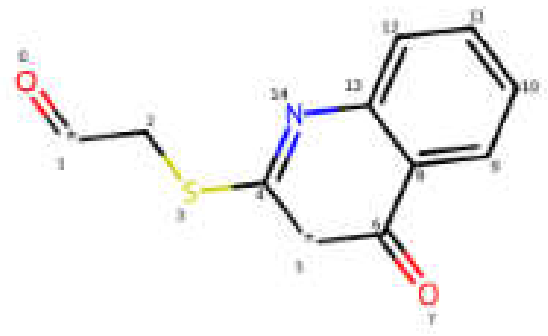

5.1.O=*CSc1:*:c(=O)c2ccccc2n1: 771

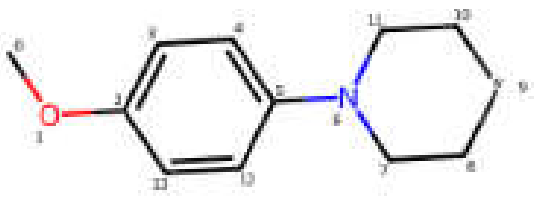

9.COc1ccc(N2CC*CC2)cc1: 753

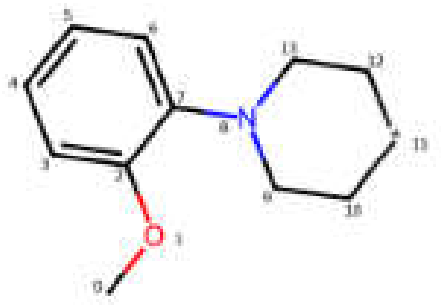

11.COc1ccccc1N1CC*CC1: 710

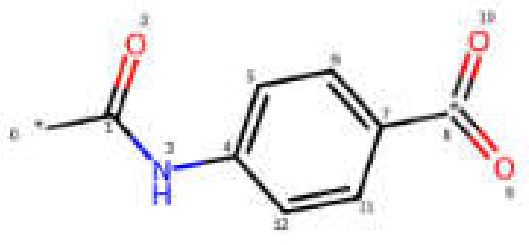

0.8.*C(=O)Nc1ccc(*(=O)=O)cc1: 681

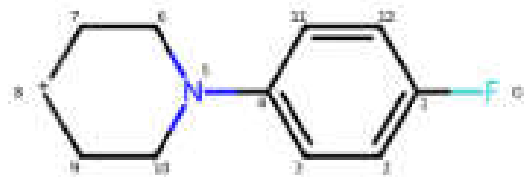

8.Fc1ccc(N2CC*CC2)cc1: 661

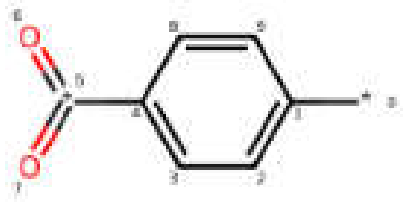

5.0.*c1ccc(*(=O)=O)cc1: 657

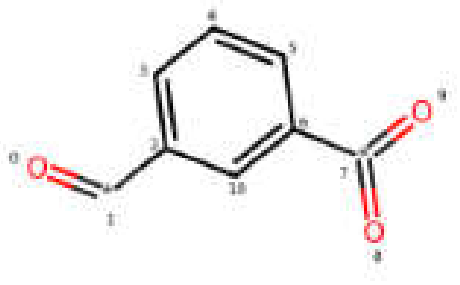

1.7.O=*c1cccc(*(=O)=O)c1: 624

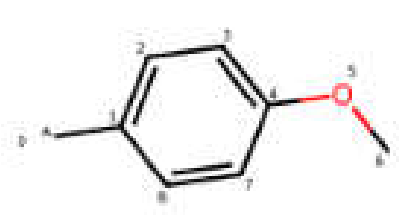

0.*c1ccc(OC)cc1: 610

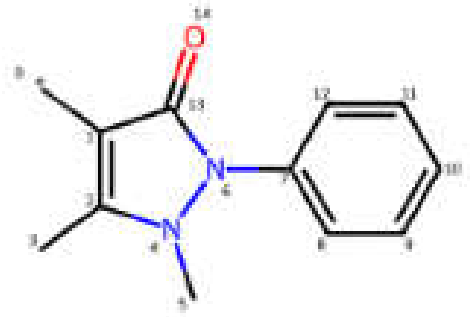

0.*c1c(C)n(C)n(-c2ccccc2)c1=O: 605

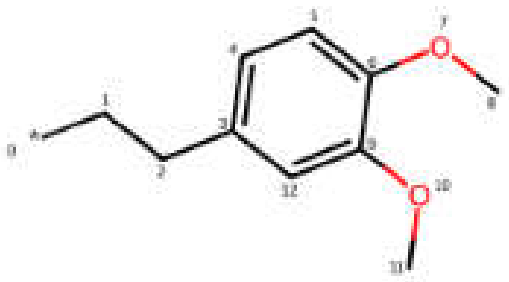

0.*CCc1ccc(OC)c(OC)c1: 598

## Linker Atom Counts

: linker atom count histogram

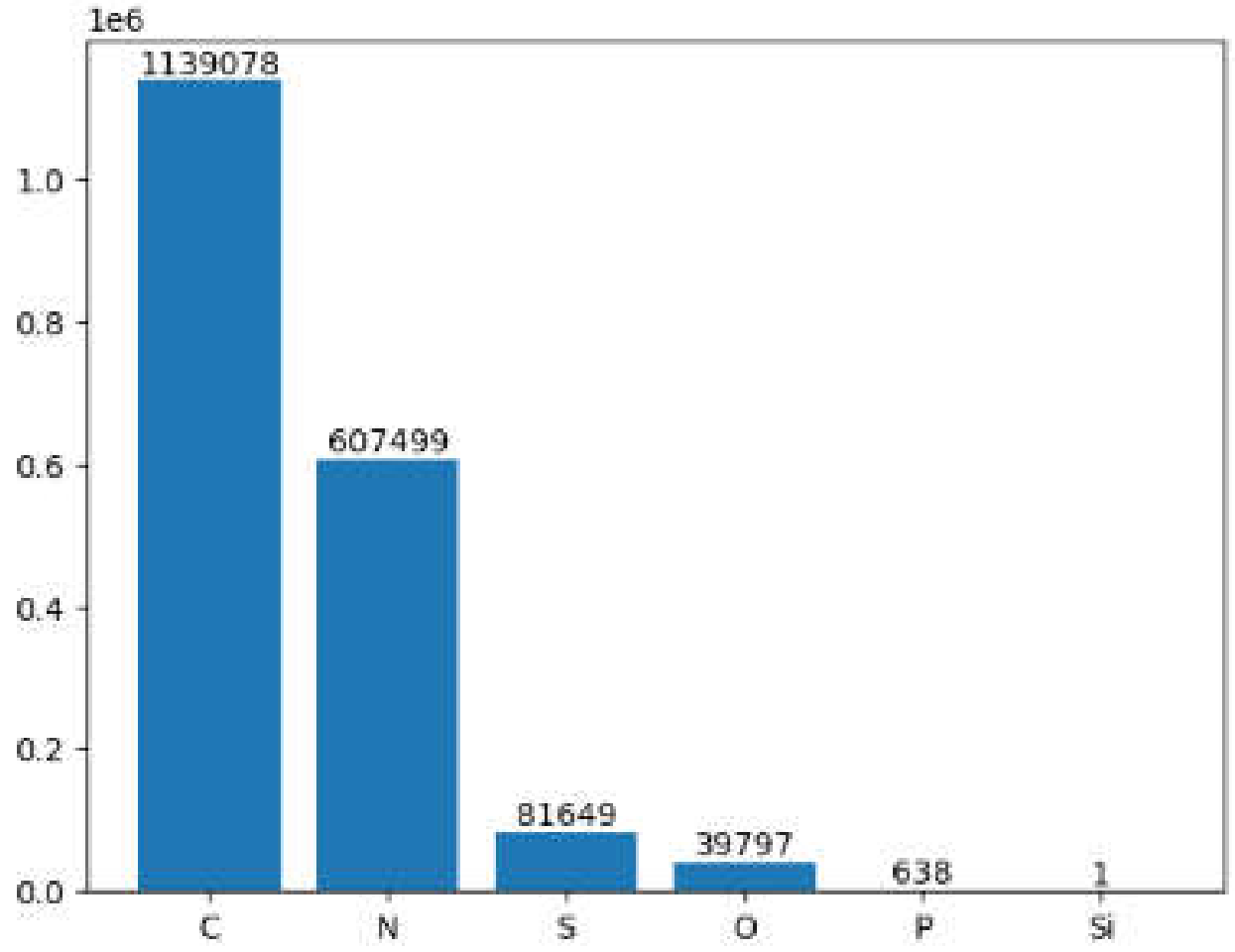

## Linker Counts

: histogram of linker counts in a molecule

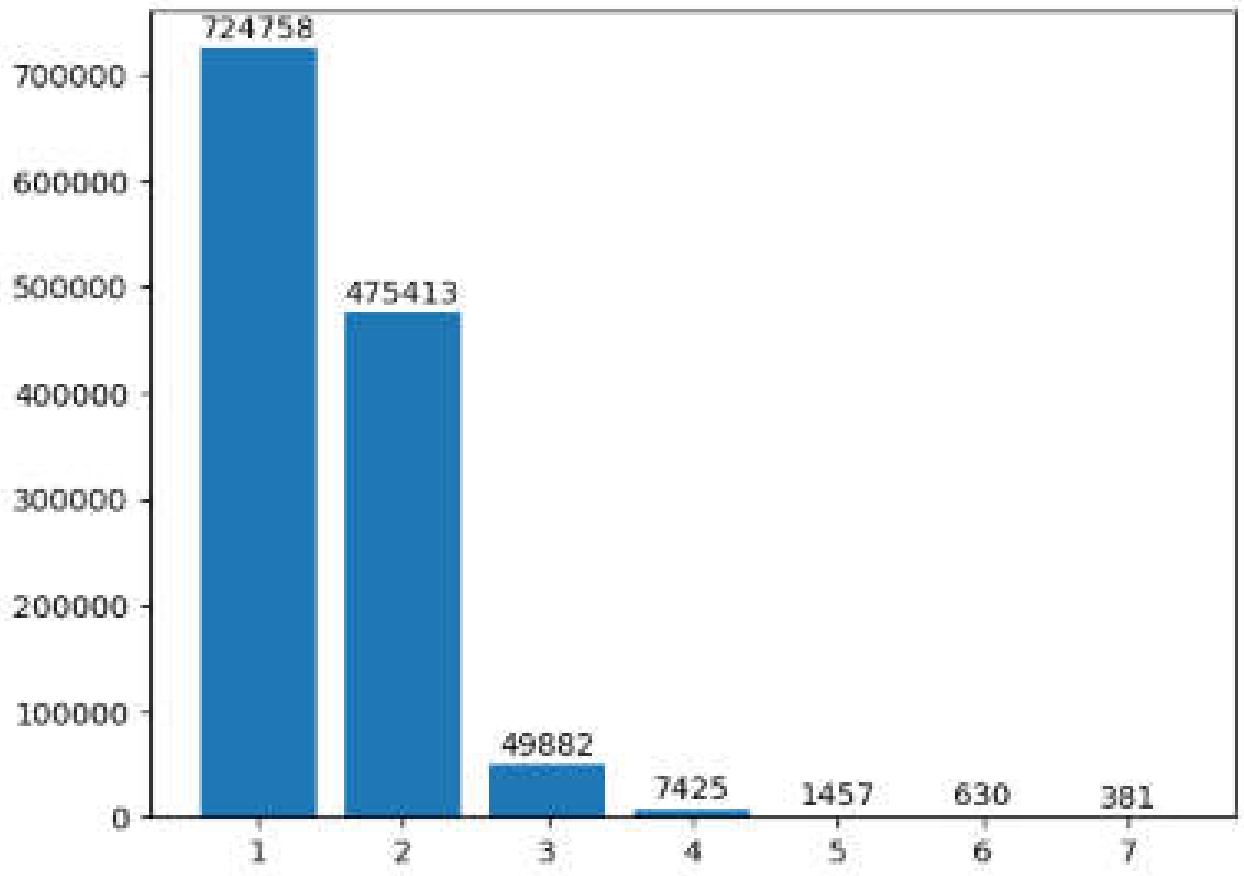

# 4. After R-Group attachment + common R-Groups filtering (reduce long tail)

(GEOM mol, core, attached R-Groups): 1,054,787 pairs

**Filtering**

- 99.99 percentile R-Groups = common R-Groups
- if more than half of total attached R-Groups common → filter out

**Top 80 Putative Cores histogram**

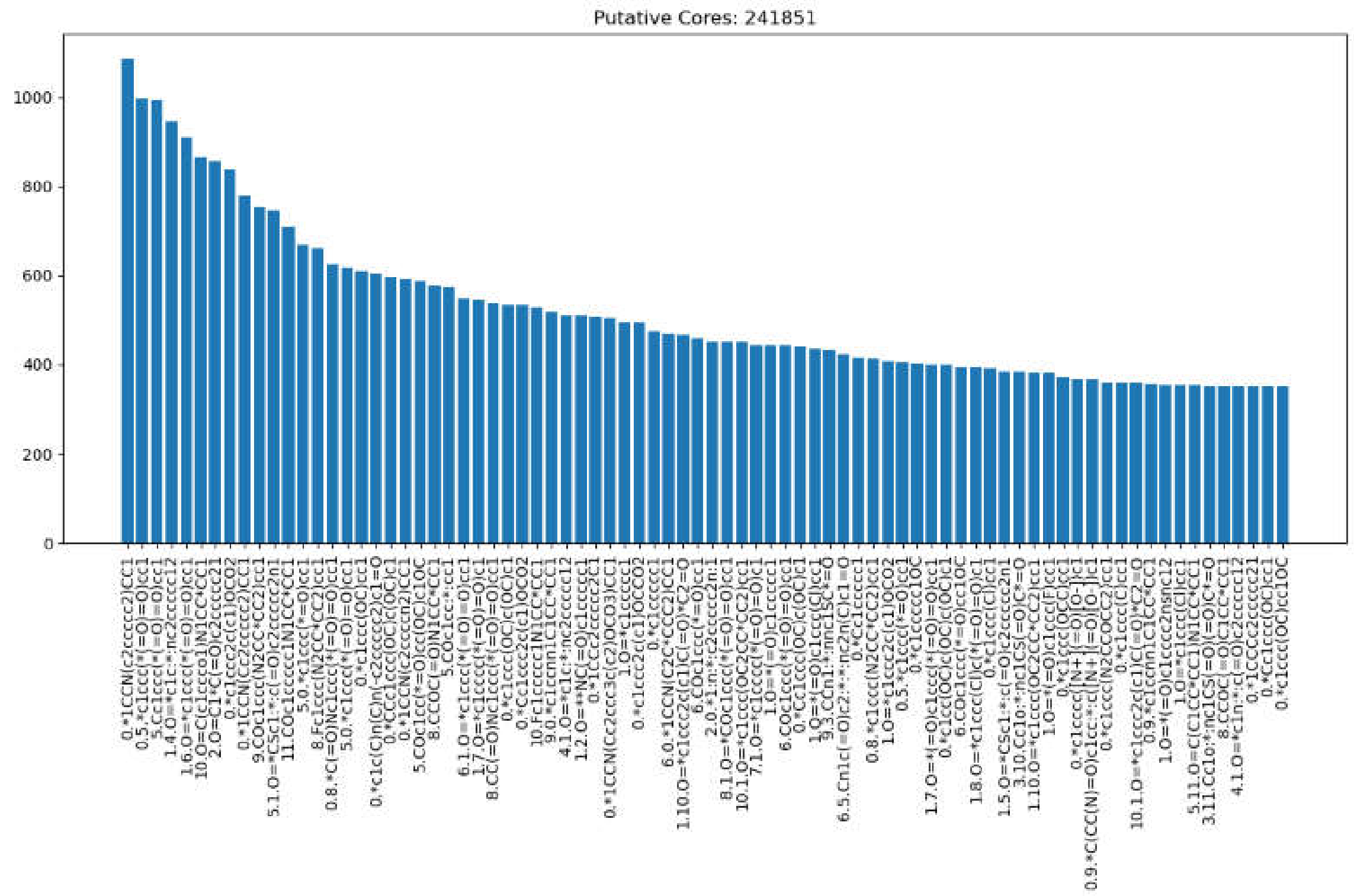

**Top 20 Putative Cores image**

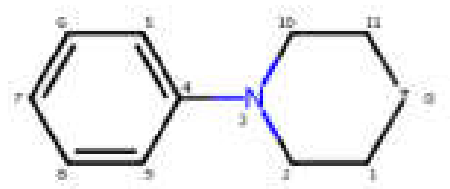

0.*1CCN(c2ccccc2)CC1: 1088

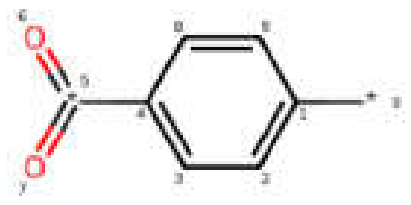

0.5.*c1ccc(*(=O)=O)cc1: 998

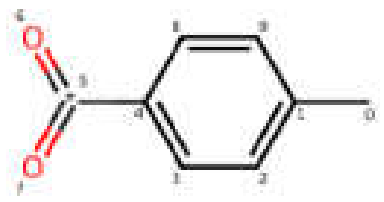

5.Cc1ccc(*(=O)=O)cc1: 994

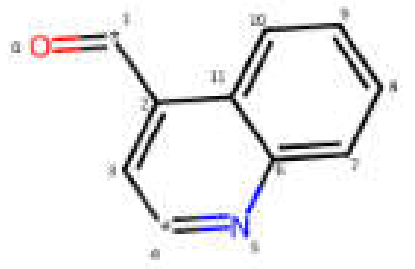

1.4.O=*c1c:*:nc2ccccc12: 946

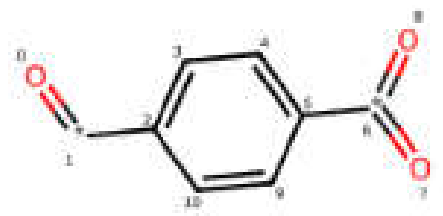

1.6.O=*c1ccc(*(=O)=O)cc1: 911

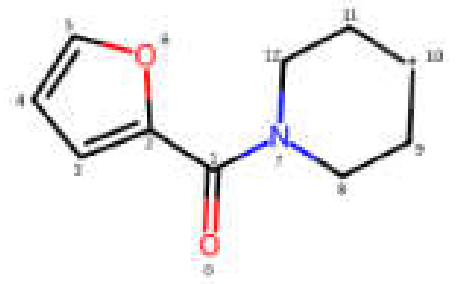

10.O=C(c1ccco1)N1CC*CC1: 867

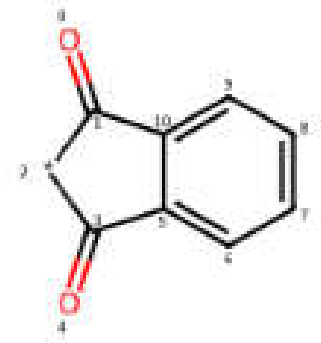

2.O=C1*C(=O)c2ccccc21: 857

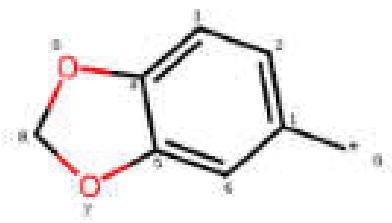

0.*c1ccc2c(c1)OCO2: 838

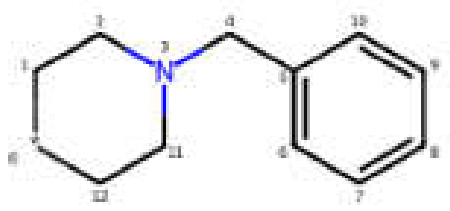

0.*1CCN(Cc2ccccc2)CC1: 779

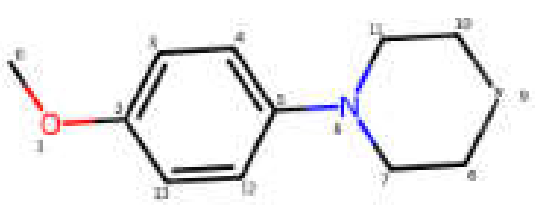

9.COc1ccc(N2CC*CC2)cc1: 753

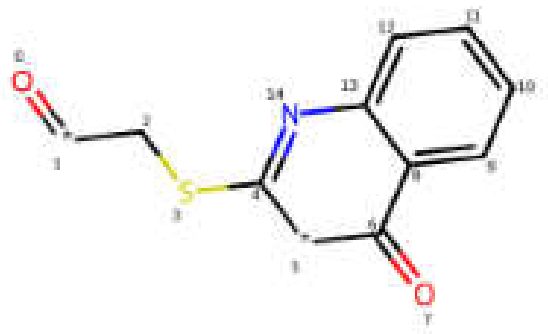

5.1.O=*CSc1:*:c(=O)c2ccccc2n1: 746

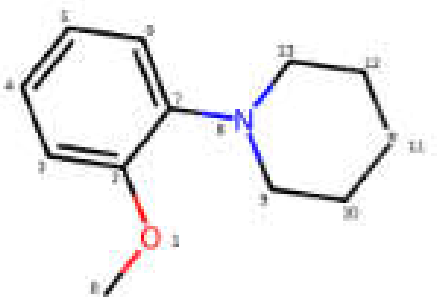

11.COc1ccccc1N1CC*CC1: 709

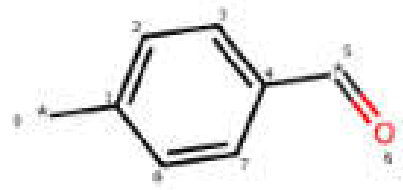

5.0.*c1ccc(*=O)cc1: 669

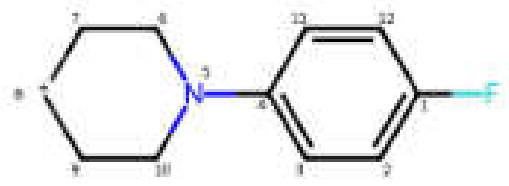

8.Fc1ccc(N2CC*CC2)cc1: 661

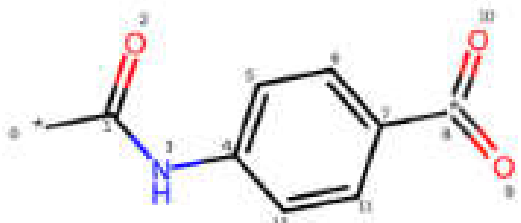

0.8.*C(=O)Nc1ccc(*(=O)=O)cc1: 626

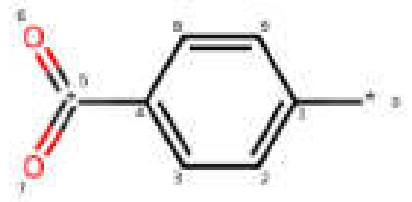

5.0.*c1ccc(*(=O)=O)cc1: 618

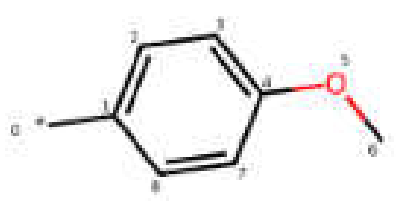

0.*c1ccc(OC)cc1: 609

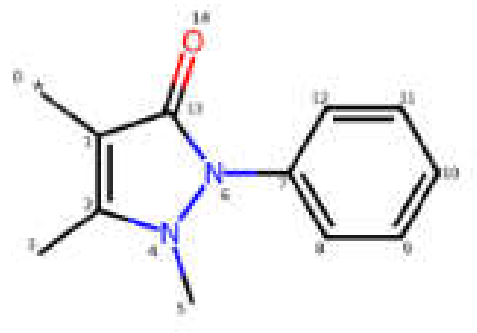

0.*c1c(C)n(C)n(-c2ccccc2)c1=O: 605

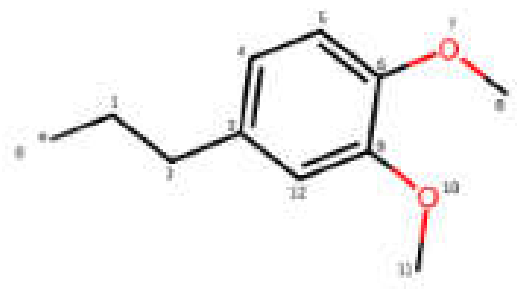

0.*CCc1ccc(OC)c(OC)c1: 598

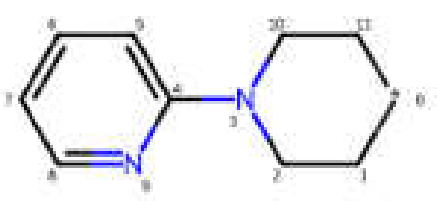

0.*1CCN(c2ccccn2)CC1: 592

**Top 80 R-Groups histogram**

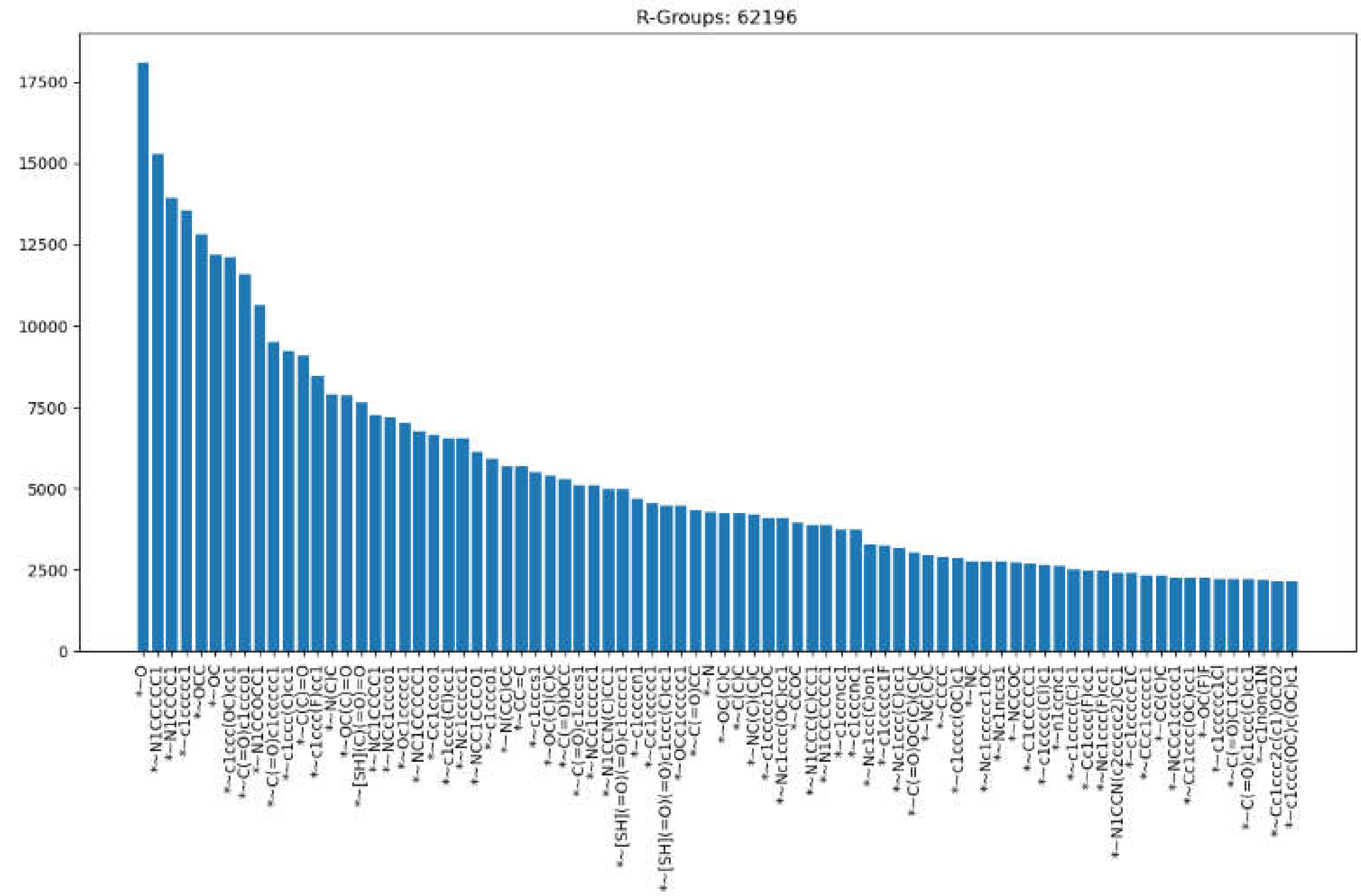

**Top 20 R-Groups image**

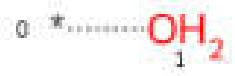

*~O: 18088

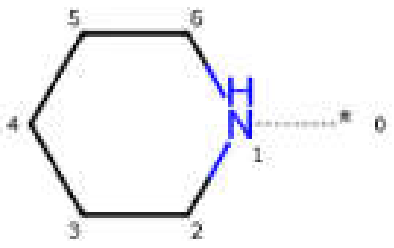

*~N1CCCCC1: 15272

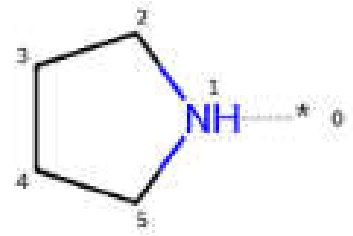

*~N1CCCC1: 13937

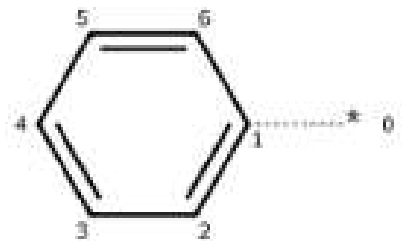

*~c1ccccc1: 13563

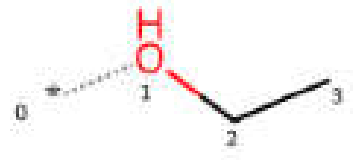

*~OCC: 12802

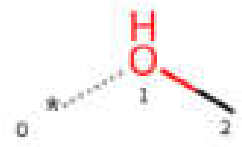

*~OC: 12189

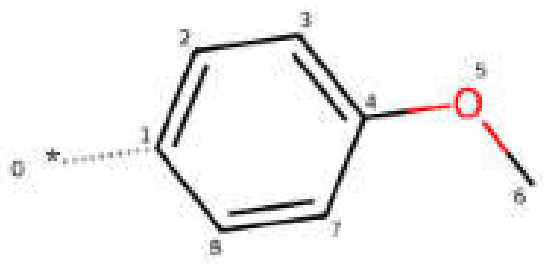

*~c1ccc(OC)cc1: 12105

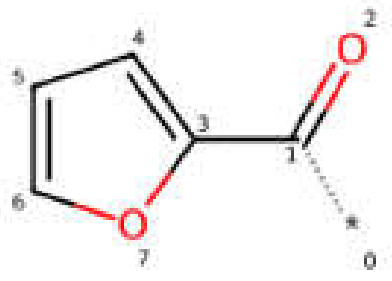

*~C(=O)c1ccco1: 11587

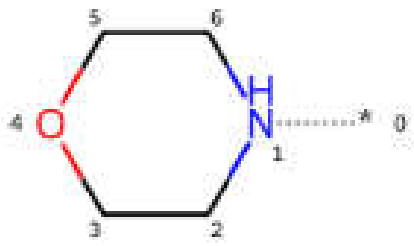

*~N1CCOCC1: 10643

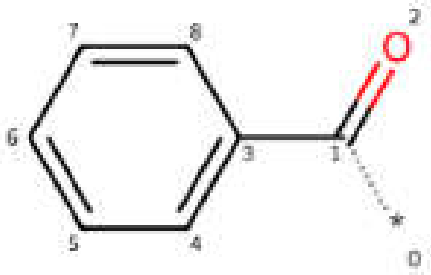

*~C(=O)c1ccccc1: 9504

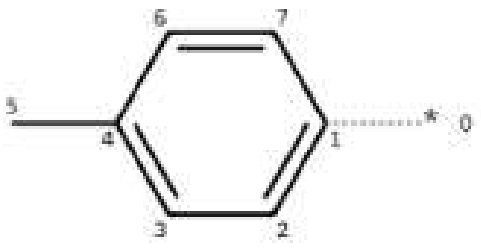

*~c1ccc(C)cc1: 9239

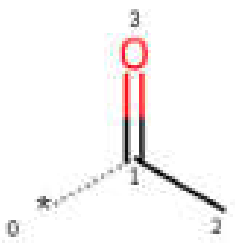

*~C(C)=O: 9098

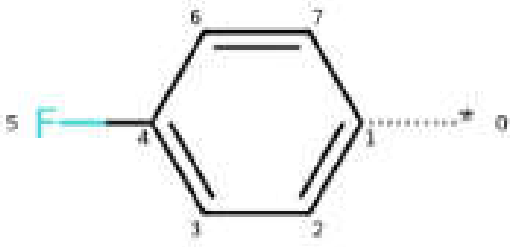

*~c1ccc(F)cc1: 8440

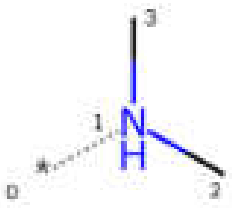

*~N(C)C: 7888

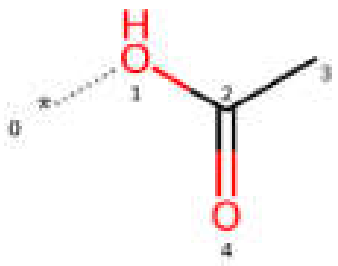

*~OC(C)=O: 7859

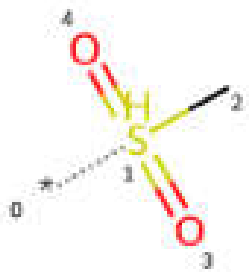

*~[SH](C)(=O)=O: 7653

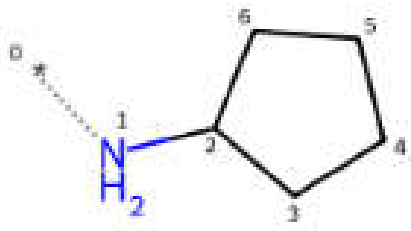

*~NC1CCCC1: 7244

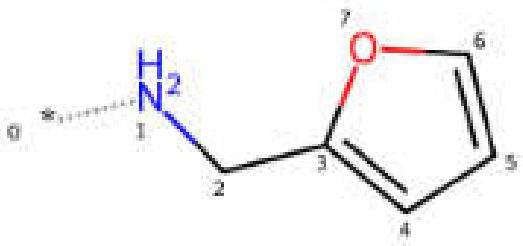

*~NCc1ccco1: 7203

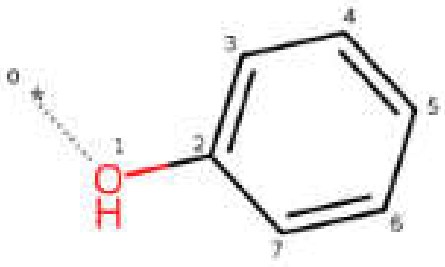

*~Oc1ccccc1: 7028

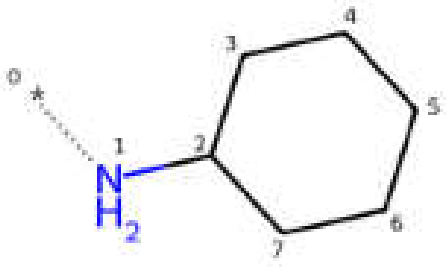

*~NC1CCCCC1: 6764

## Compare Putative Cores

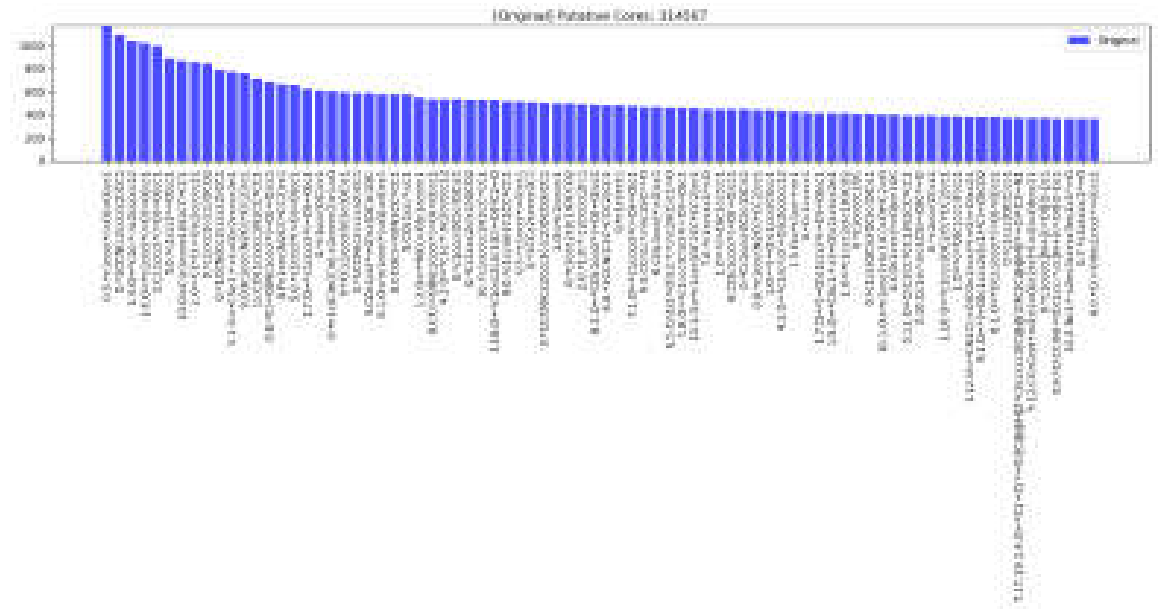

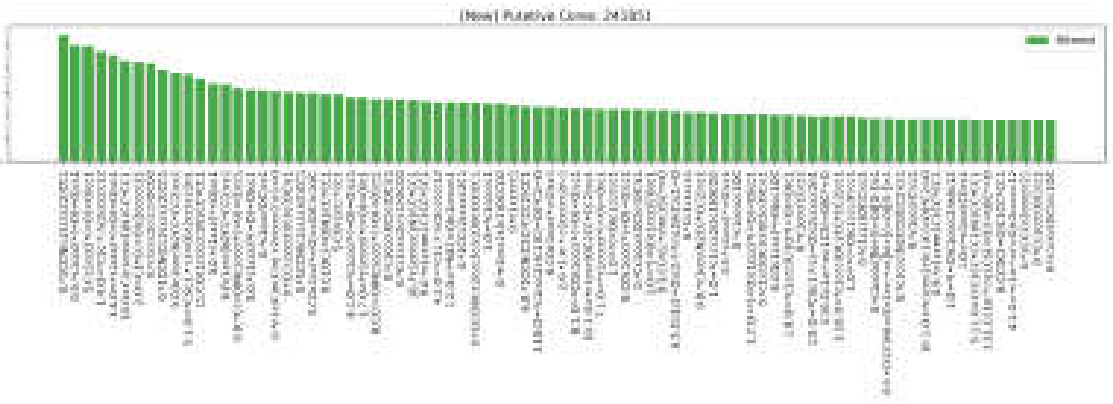

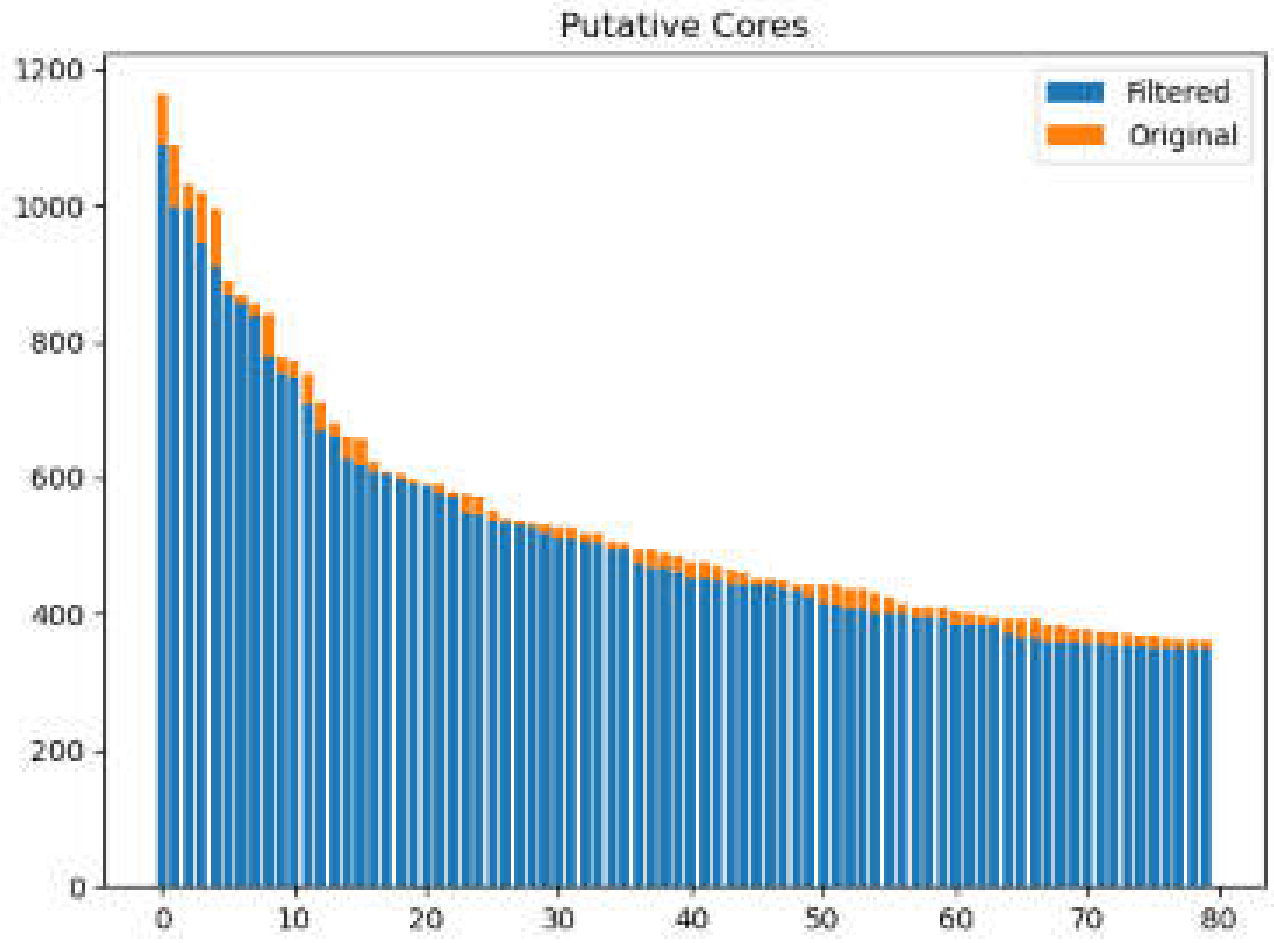

## Compare R-Groups

⇒ reduced long tail

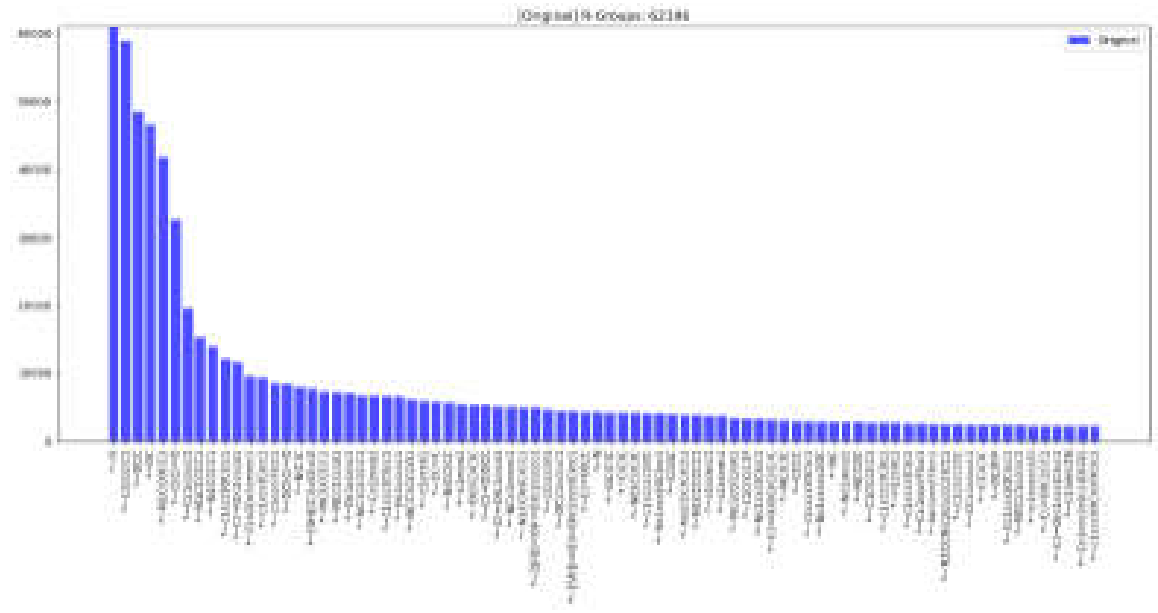

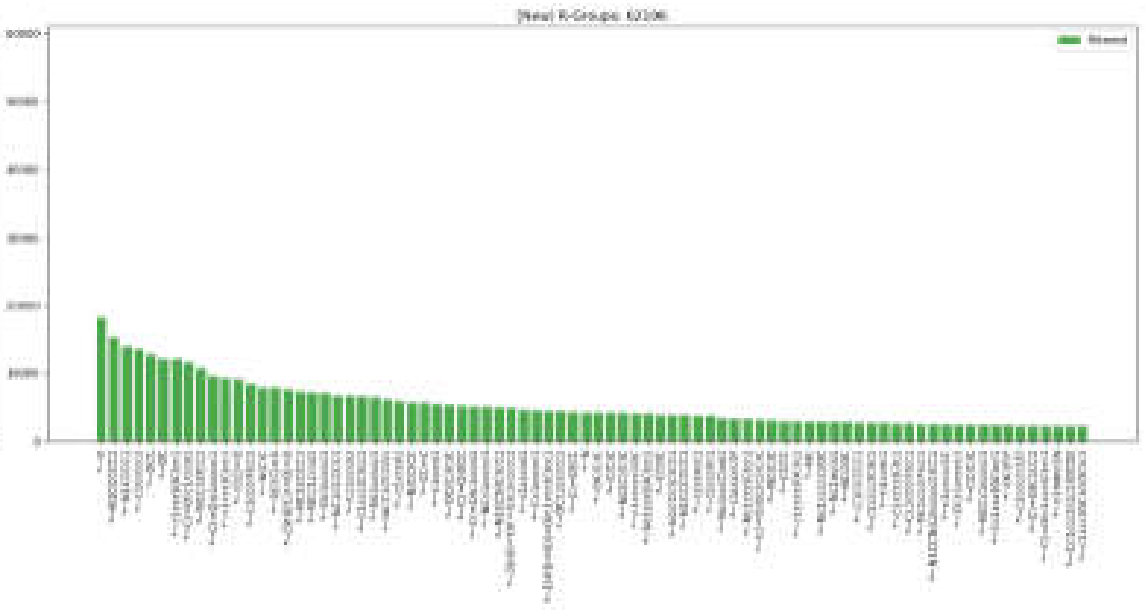

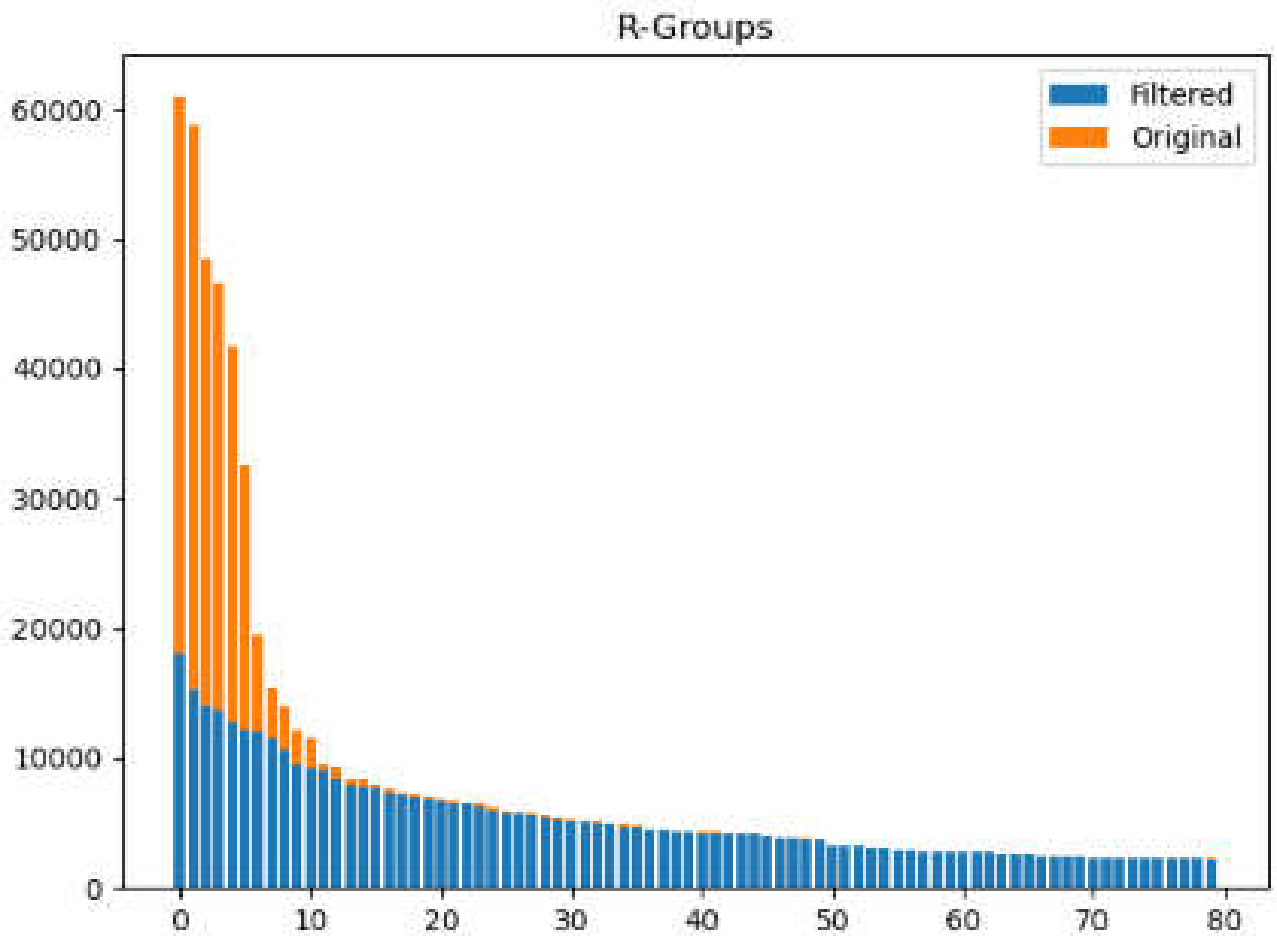

**S2:** Comparison between two molecular graph decomposition methods using Mucko Scaffolds and Naveja's Putative Core framework

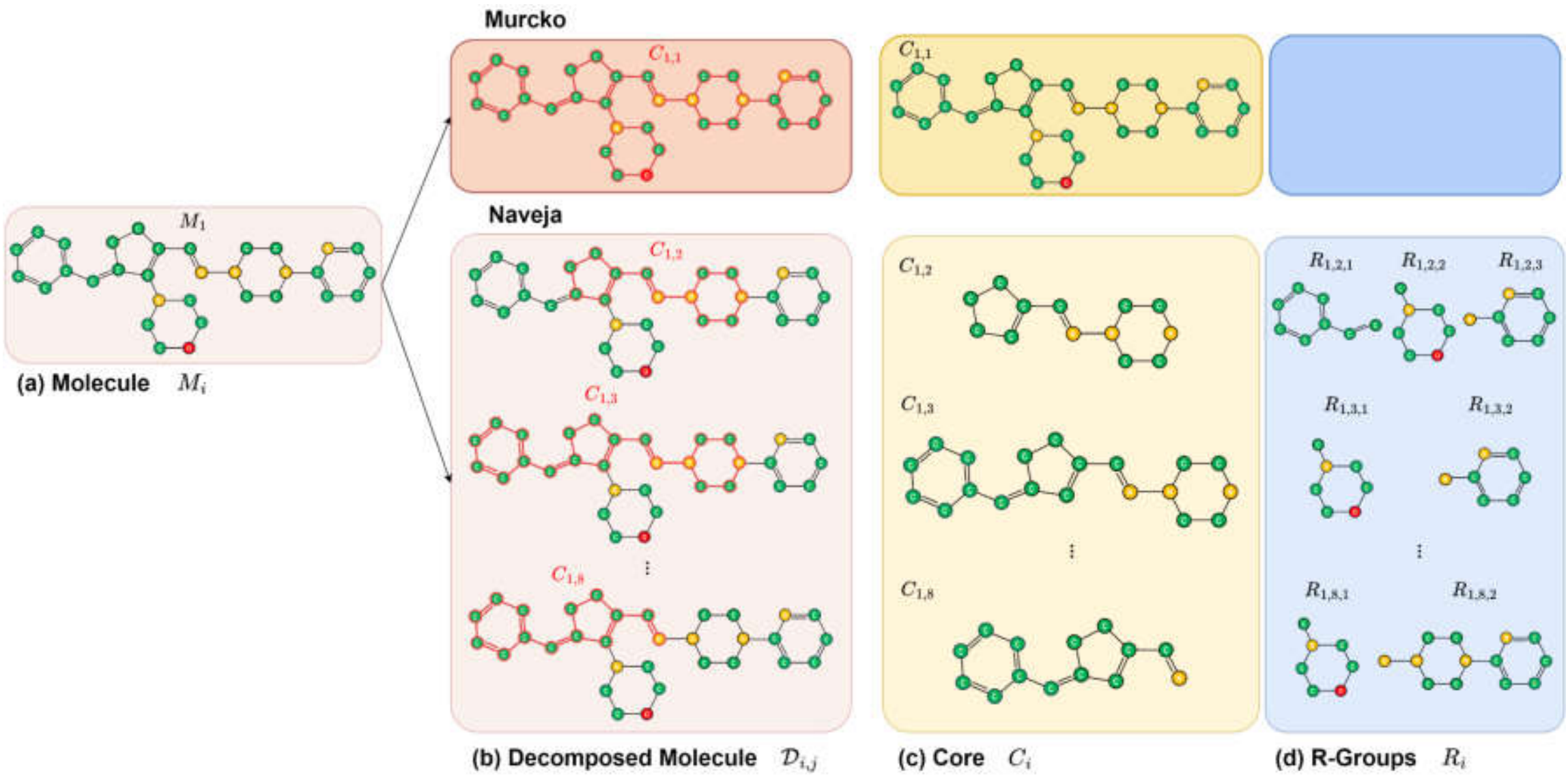

**S3:** Preprocessing Pipeline for the MolPLA pre-training dataset

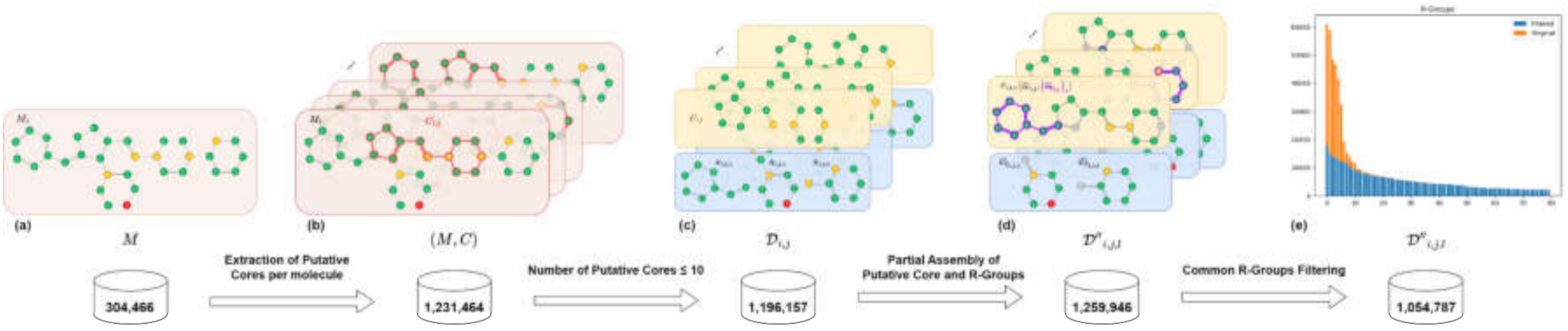

**S4:** Detailed illustration for our proposed molecular graph decomposition

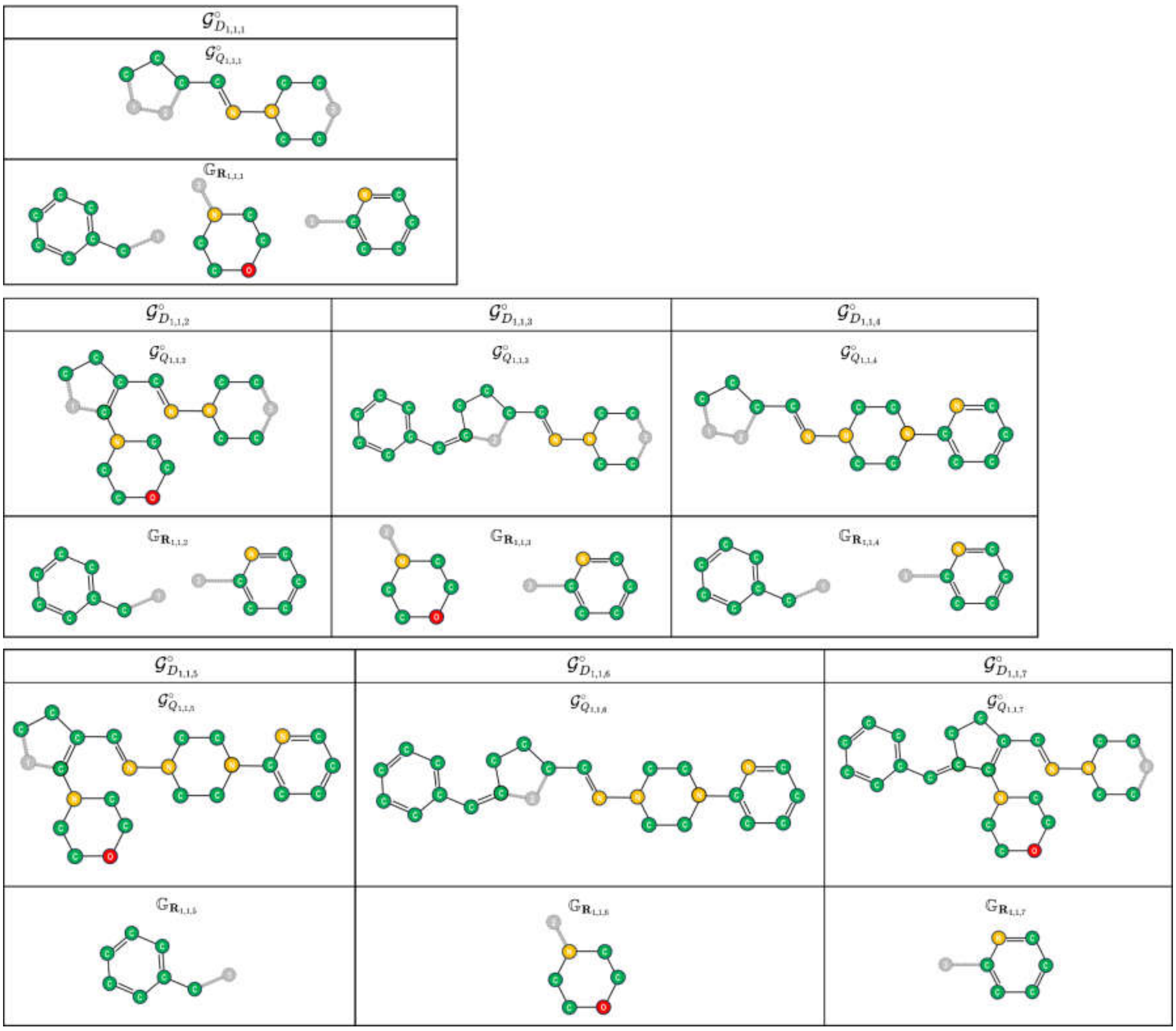

## **S5:** Implementation details for MolPLA

The weight-sharing GNN-based **graph encoder** $f_\theta$ used in both MolPLA's Masked Graph Contrastive Learning and R-Group Retrieval Framework consists of five GIN layers followed by Layer Normalization applied to the updated node representations. The initial node and edge representations from graph input $\mathcal{G}$ are a summation of attribute-wise embeddings based on predefined molecular information obtained from Rdkit which are the following,

| **Attribute Type** | **Attribute Name** | **Number of Features** |
|---|---|---|
| Node (Atom Features) | Atomic Number | 128 |
| | Formal Charge | 11 |
| | Chirality Tag | 9 |
| | Hybridization | 9 |
| | Number of Explicit Hs | 9 |
| | Aromaticity | 2 |
| Edge (Bond Features) | Conjugation | 2 |
| | Bond Type | 22 |
| | Bond Direction | 7 |
| | Bond Stereochemistry | 6 |
| | Aromaticity | 2 |

The GIN layer updates node representations based on its neighbor information. The update rule for node $v$ in the $l$th layer is mathematically expressed as follows,

$$h_v^{(l+1)} = \mathrm{MLP}^{(l)} \left( (1 + \epsilon^{(l)}) \cdot h_v^{(l)} + \sum_{u \in \mathcal{N}(v)} h_u^{(l)} \right)$$

where $h_v^{(l)}$ is the feature vector of node $v$ at layer $l$, $\mathcal{N}(v)$ denotes the set of neighbors of $v$, $\mathrm{MLP}^{(l)}$ is a multi-layer perceptron at layer $l$, and $\epsilon^{(l)}$ is a learnable parameter.

The updated node representations from each GIN layer are sequentially applied by layer normalization, ReLU activation (except the last) and dropout. The lastly updated node representations are the final output from $f_\theta$ which are expressed as $H_{\mathcal{G}_\mathrm{M}}$ and $H_{\mathcal{G}_\mathrm{M}}$ for the original and decomposed molecular graph respectively.

**S6:** Illustration of the MolPLA's functionality in lead optimization

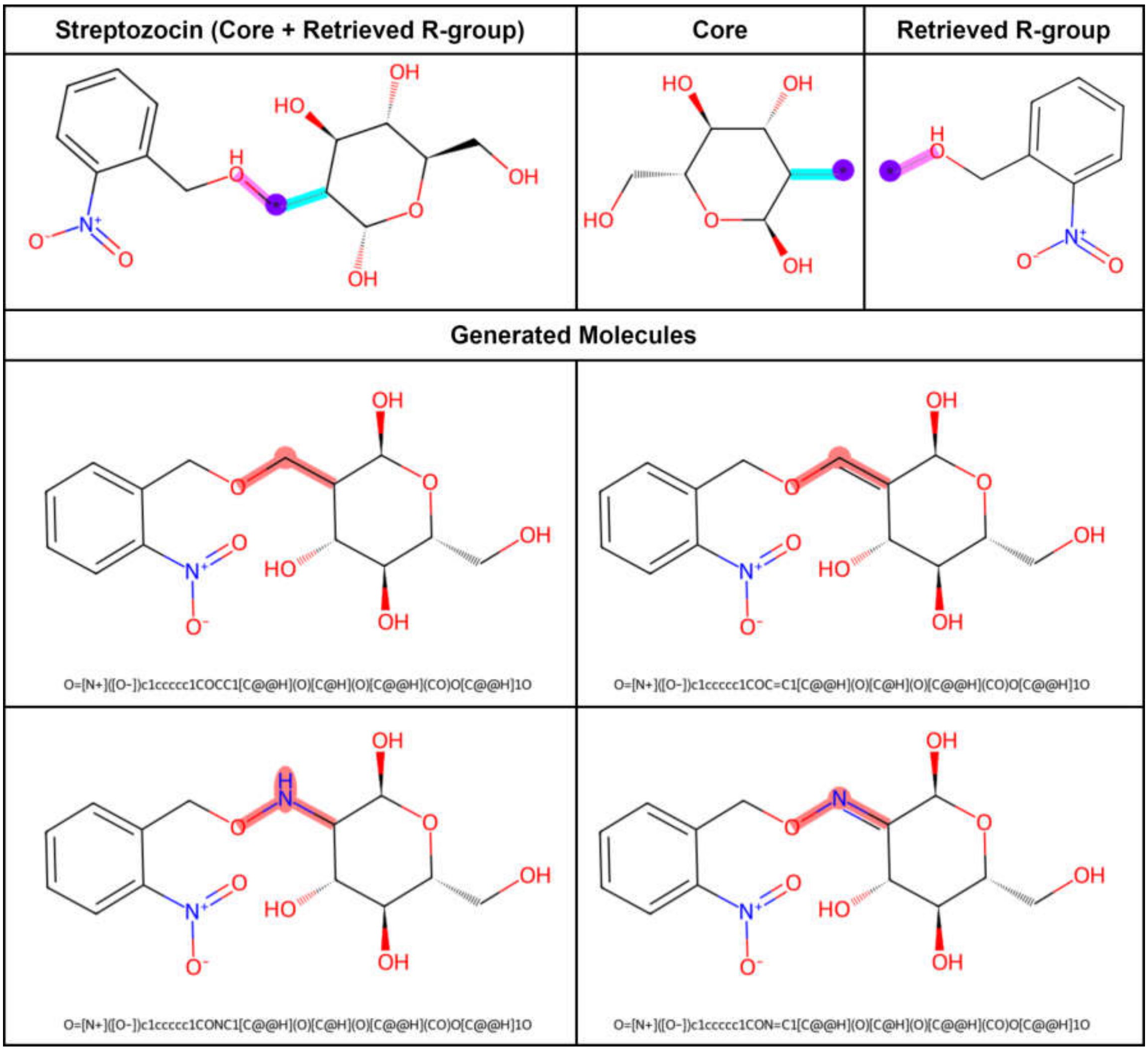

The linker node is first designated based on the node coloring results from the node representations of the query template (**core**) graph applied with PCA. The core graph containing the linker node with all its node-edge attributes masked is fed to MolPLA's pre-trained graph encoder. Among the encoded node representations, the linker node is exclusively selected, concatenated with the R-group condition vector and fed to MolPLA's query linker node projection head to **retrieve top nearest R-groups** in co-embedding space. For each R-group with its masked linker joint being re-attached to the query template, we invoke the enumeration process of all possible atom and bond features being used to "fill in" the masked attributes, and obtain re-attached molecules that are valid. These re-attached molecules are deemed as **generated molecules** and there may be multiple results since different combinations of unmasking atom and bond features can satisfy chemical rules.

**S7:** Visualization of Node Representations for Riluzole and Lasmiditan

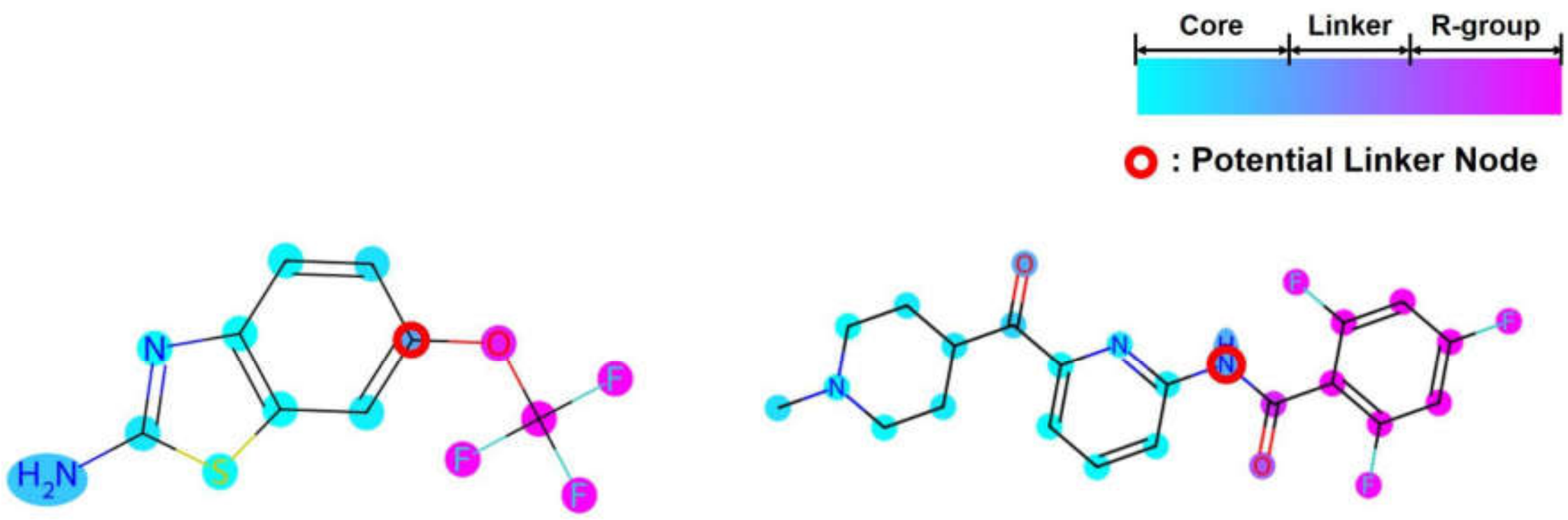

**Riluzole** **Lasmiditan**

**S8:** Distribution of QED and SAScores of molecule generated by MolPLA for Riluzole and Lasmiditan

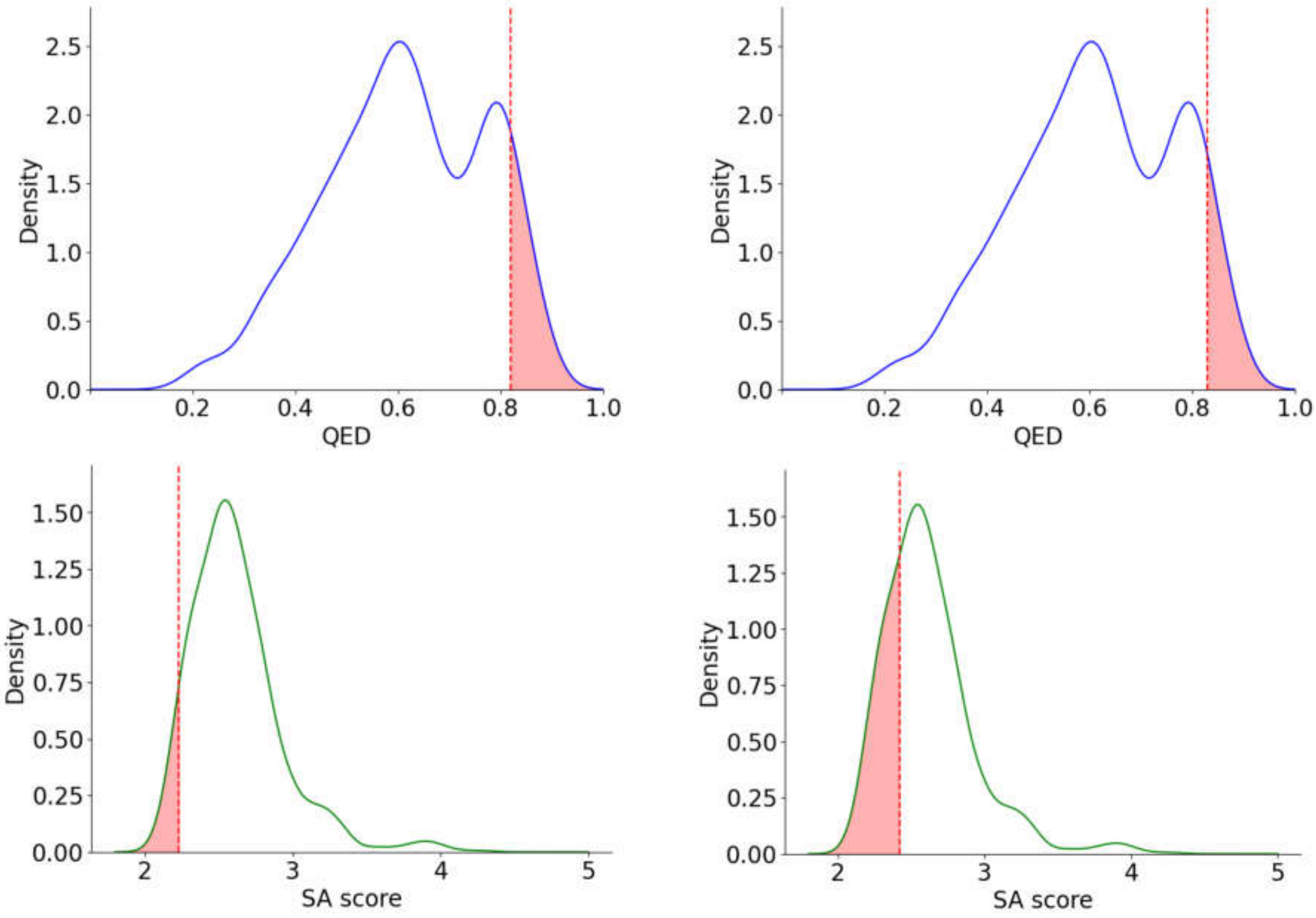

**Riluzole** **Lasmiditan**

**S9:** List of generated molecules for reference molecules Riluzole and Lasmiditan from MolPLA deployed in lead optimization scenario

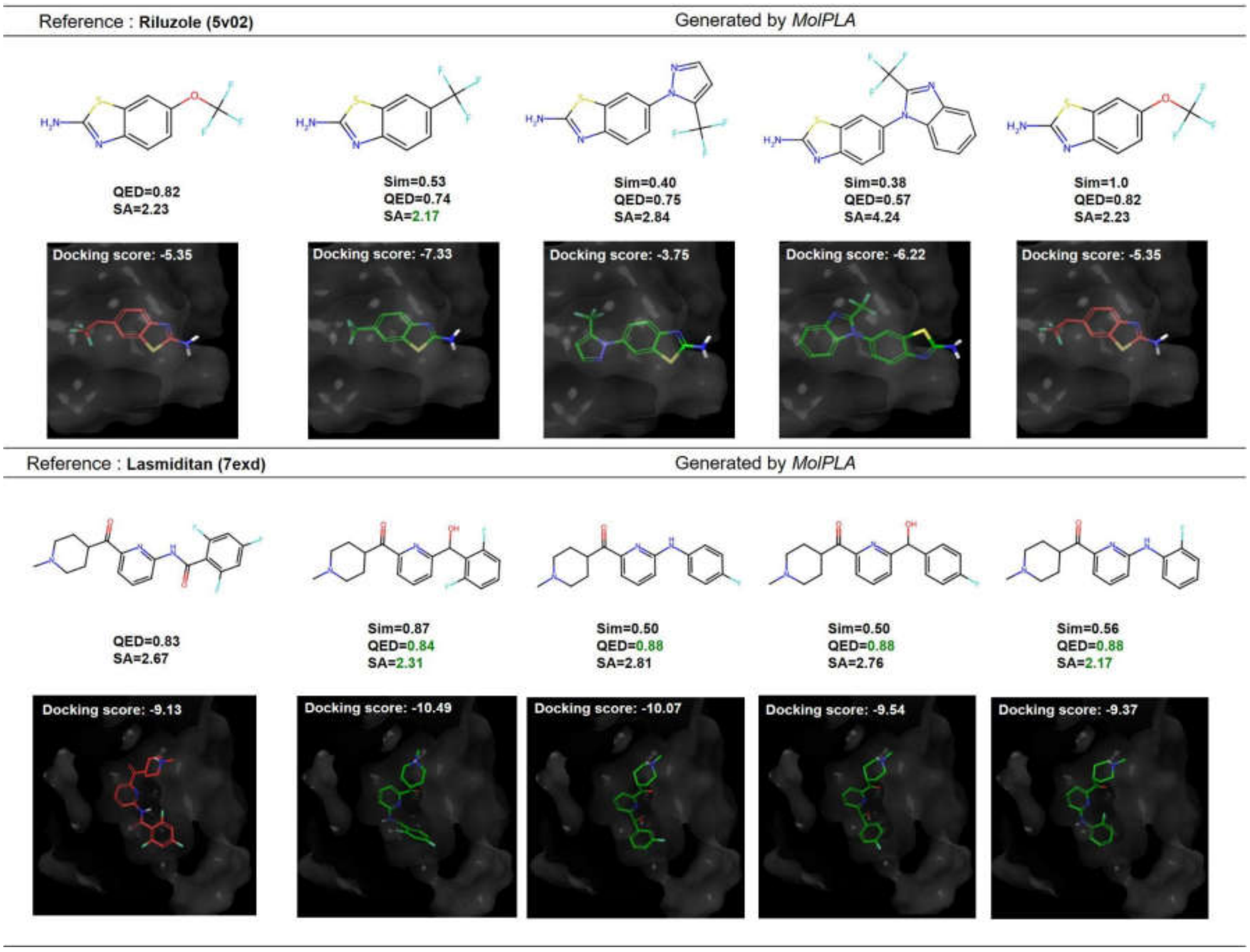

**S10:** Cross reference (Streptozocin, Capmatinib, Riluzole and Lasmiditan) analysis results for MolPLA's R-group Retrieval framework.

| Query Template | R-group Substitution | Top 5 Retrieved R-groups (SMILES) |
|---|---|---|
| Streptozocin | Streptozocin | *~n1oc(=O)[nH]c1=O, *~n1ccc(=O)[nH]c1=O, *~n1ncc(=O)[nH]c1=O, *~n1c(=O)[nH]c(=O)n(C)c1=O, *~OCc1ccccc1[N+](=O)[O-] |
| | Capmatinib | *~OCc1ccccc1F, *~OCc1ccc(F)cc1, *~OCc1c(F)cccc1F, *~NOC(=O)c1ccccc1F, *~OCc1cc(F)ccc1-c1ccc(C)cc1 |
| | Riluzole | *~OCc1c(F)cccc1F, *~NOC(=O)c1ccccc1F, *~OCc1cc(C(F)(F)F)cc(C(F)(F)F)c1, *~O[C@@H](C)c1cc(C(F)(F)F)cc(C(F)(F)F)c1, *~OC(C)c1cc(C(F)(F)F)cc(C(F)(F)F)c1 |
| | Lasmiditan | *~OCc1ccccc1F, *~OCc1ccc(F)cc1, *~OCc1c(F)cccc1F, *~NOC(=O)c1ccccc1F, *~OCc1cc(F)ccc1-c1ccc(C)cc1 |
| Capmatinib | Streptozocin | *~n1nnc(C(=O)/C=C/N(C)C)c1C, *~n1nc(C)c(C#N)c1N, *~n1nc(C)c(CN)c1C, *~n1nc(C)c(NC(=O)c2ccnn2C)c1C, *~n1nc(C)c(NC(=O)c2ccn(CC)n2)c1C |
| | Capmatinib | *~c1ccc(F)c(C(F)(F)F)c1, *~c1ccc(C(F)(F)F)c(F)c1, *~c1cccc(C(F)(F)F)c1, *~c1ccc(F)c2ccccc21, *~c1cc(C(F)(F)F)cc(C(F)(F)F)c1 |
| | Riluzole | *~n1nc(C(F)F)cc1C(F)F, *~n1nc(C(F)(F)F)cc1C(F)(F)F, *~c1cc(C(F)(F)F)cc(C(F)(F)F)c1, *~n1c(C(F)(F)F)nc2cc(C(F)(F)F)ccc21, *~n1c(-c2ccc(F)cc2)nc2cc(C(F)(F)F)ccc21 |
| | Lasmiditan | *~c1ccc(F)c(C(F)(F)F)c1, *~c1ccc(C(F)(F)F)c(F)c1, *~c1cccc(C(F)(F)F)c1, *~c1ccc(F)c2ccccc21, *~c1cc(C(F)(F)F)cc(C(F)(F)F)c1 |
| Riluzole | Streptozocin | *~n1ccc(=O)cc1, *~n1nc(C)c[n+]1[O-], *~n1cccc1/C=N\O, *~n1cccc1/C=N/O, *~n1c(C)cc(/C=N/O)c1C |
| | Capmatinib | *~[NH2+]c1ccc(-c2ccccc2F)nn1, *~Nc1ncc(C(F)(F)F)c(NC2CCC2)n1, *~Nc1nc(-c2ccco2)cc(C(F)(F)F)n1, *~[NH2+]c1ncc(F)c(-c2cc(F)c3nc(C)n(C(C)C)c3c2)n1 *~Nc1ncc(F)c(-c2cc(F)c3nc(C)n(C(C)C)c3c2)n1 |
| | Riluzole | *~c1c(F)c(F)c(C(F)(F)F)c(F)c1F, *~Oc1c(F)c(F)c(C(F)(F)F)c(F)c1F, *~n1c(C(F)(F)F)nc2cc(C(F)(F)F)ccc21, *~n1c(-c2ccc(F)cc2)nc2cc(C(F)(F)F)ccc21, *~[NH2+]c1ncc(F)c(-c2cc(F)c3nc(C)n(C(C)C)c3c2)n1 |
| | Lasmiditan | *~[NH2+]c1ccc(-c2ccccc2F)nn1, *~Nc1ncc(C(F)(F)F)c(NC2CCC2)n1, *~Nc1nc(-c2ccco2)cc(C(F)(F)F)n1, *~[NH2+]c1ncc(F)c(-c2cc(F)c3nc(C)n(C(C)C)c3c2)n1 *~Nc1ncc(F)c(-c2cc(F)c3nc(C)n(C(C)C)c3c2)n1 |
| Lasmiditan | Streptozocin | *~n1cc(C#N)c2c(N)ncnc21, *~n1c[n+](C)c2cc(C)c(C)cc21, *~n1c([NH3+])nc2ccccc21, *~n1cc[n+](CC=C)c1, *~n1cc([NH3+])cn1 |
| | Capmatinib | *~[NH2+]c1ncc(-c2ccc(F)cc2)n1C, *~Nc1ncc(-c2ccc(F)cc2)n1C, *~Nc1nc2ccccc2n1Cc1ccc(F)cc1, *~[NH2+]c1ncc(F)c(-c2cc(F)c3nc(C)n(C(C)C)c3c2)n1, *~Nc1nc2c(c(=O)n(C)c(=O)n2C)n1Cc1ccc(F)cc1 |
| | Riluzole | *~n1c(-c2ccc(F)cc2)noc1=O, *~n1cc(C(=O)O)c(=O)c2cc(F)c(F)c(F)c21, *~n1c(C(F)(F)F)nc2cc(C(F)(F)F)ccc21, *~n1c(-c2ccc(F)cc2)nc2cc(C(F)(F)F)ccc21, *~n1nnc(-c2cc(C(F)(F)F)cc(C(F)(F)F)c2)n1 |
| | Lasmiditan | *~[NH2+]c1ncc(-c2ccc(F)cc2)n1C, *~Nc1ncc(-c2ccc(F)cc2)n1C, *~Nc1nc2ccccc2n1Cc1ccc(F)cc1, *~[NH2+]c1ncc(F)c(-c2cc(F)c3nc(C)n(C(C)C)c3c2)n1, *~Nc1nc2c(c(=O)n(C)c(=O)n2C)n1Cc1ccc(F)cc1 |